\documentclass{article}

\usepackage{arxiv}

\usepackage[utf8]{inputenc}
\usepackage[T1]{fontenc}
\usepackage{hyperref}
\usepackage{url}
\usepackage{booktabs}
\usepackage{nicefrac}
\usepackage{microtype}
\usepackage{graphicx}

\usepackage[square,numbers,sort&compress]{natbib}

\usepackage{doi}
\usepackage{multirow}

\usepackage{amsmath,amssymb,amsfonts}
\usepackage{amsthm}
\usepackage{mathrsfs}
\usepackage{bm}
\usepackage{gensymb}

\usepackage{cleveref}

\usepackage{algorithm}
\usepackage{algorithmicx}
\usepackage{algpseudocode}

\usepackage{listings}
\usepackage{xcolor}
\usepackage{tabularx}
\usepackage{enumerate}
\usepackage{wrapfig}
\usepackage{adjustbox}
\usepackage{bbding}
\usepackage{array}
\usepackage{ragged2e}

\newcolumntype{L}[1]{>{\RaggedRight\arraybackslash}p{#1}}
\newcolumntype{C}[1]{>{\centering\arraybackslash}p{#1}}

\theoremstyle{plain}

\theoremstyle{definition}

\title{Synergistic Event-SVE Imaging for Quantitative Propellant Combustion Diagnostics}

\date{}

\newif\ifuniqueAffiliation
\uniqueAffiliationtrue

\usepackage{authblk}

\author[1,2,$\dagger$]{Jing Tao}
\author[1,2,$\dagger$]{Taihang Lei}
\author[1,2,*]{Banglei Guan}
\author[1,3]{Ying Qu}
\author[1,3]{Xudong Na}
\author[1,3]{Likun Ma}
\author[1,2]{Yang Shang}
\author[1,2]{Qifeng Yu}

\affil[1]{College of Aerospace Science and Engineering, National University of Defense Technology, Changsha 410073, China}
\affil[2]{Hunan Provincial Key Laboratory of Image Measurement and Vision Navigation, National University of Defense Technology, Changsha 410073, China}
\affil[3]{Hypersonic Technology Laboratory, National University of Defense Technology, Changsha 410073, China}
\affil[*]{Corresponding author: guanbanglei12@nudt.edu.cn}
\affil[$\dagger$]{The two authors contribute equally to this work.}

\renewcommand{\shorttitle}{Synergistic Event-SVE Imaging}

\hypersetup{
	pdftitle={Synergistic Event-SVE Imaging for Quantitative Propellant Combustion Diagnostics},
	pdfauthor={Jing Tao, Taihang Lei, Banglei Guan et al.},
	pdfkeywords={Event camera, SVE camera, HDR imaging, Propellant combustion, Particle metrology},
	colorlinks=true,
	linkcolor=blue,
	citecolor=blue,
	urlcolor=blue,
	anchorcolor=blue
}

\begin{document}
	\maketitle
	
	\begin{abstract}
		Real-time monitoring of high-energy propellant combustion is difficult. Extreme high dynamic range (HDR), microsecond-scale particle motion, and heavy smoke often occur together. These conditions drive saturation, motion blur, and unstable particle extraction in conventional imaging.
		We present a closed-loop Event--SVE measurement system that couples a spatially variant exposure (SVE) camera with a stereo pair of neuromorphic event cameras. The SVE branch produces HDR maps with an explicit smoke-aware fusion strategy. A multi-cue smoke-likelihood map is used to separate particle emission from smoke scattering, yielding calibrated intensity maps for downstream analysis.
		The resulting HDR maps also provide the absolute-intensity reference missing in event cameras. This reference is used to suppress smoke-driven event artifacts and to improve particle-state discrimination. Based on the cleaned event observations, a stereo event-based 3D pipeline estimates separation height and equivalent particle size through feature extraction and triangulation (maximum calibration error 0.56\%).
		Experiments on boron-based propellants show multimodal equivalent-radius statistics. The system also captures fast separation transients that are difficult to observe with conventional sensors. Overall, the proposed framework provides a practical, calibration-consistent route to microsecond-resolved 3D combustion measurement under smoke-obscured HDR conditions.
	\end{abstract}
	
	
	\section{Introduction}\label{sec1}
	High-energy solid propellant combustion involves complex multiphase reactions. Such reactions simultaneously produce fast-evolving particles and optically dense smoke within the same field of view. Boron powder, a common metallic additive, can form coral-like aggregates during combustion \cite{LIU2020105595,LI2023108126,Tan2025OptimalPG}. In solid-fuel ramjet engines, primary combustion in the gas generator supplies reactants for secondary combustion in the chamber. The evolution and transport of boron-based aggregates govern the particle-size distribution entering the secondary stage, making their equivalent size and separation height the primary measurands. These parameters are directly related to combustion efficiency, propellant performance, and engine stability \cite{SALGANSKY,PATEL}. High-precision in-situ monitoring is an important engineering objective, supporting formulation optimization, instability suppression, and improved engine reliability.
	
	In practice, existing diagnostic techniques face a trade-off among temporal resolution, spatial resolution, and dynamic range. Recent studies describe boron agglomeration on the burning surface as a multistage process, including accumulation, coalescence, growth, and eventual detachment \cite{RASHKOVSKIY2019277,Rashkovskiy03082017,Rashkovskiy_2018}. These stages are strongly influenced by propellant formulation and combustion pressure \cite{LI2023108126,Rashkovskiy03082017,guan2026fusion}.
	Capturing these dynamics requires microsecond temporal resolution for motion, sufficient spatial resolution for morphology, and extreme HDR imaging to observe incandescent particles against a smoke-filled background.
	The central difficulty is that these conditions are coupled: intensely luminous, fast-moving particles coexist with dense, time-varying smoke that degrades visibility and destabilizes feature extraction.
	
	Traditional high-speed cameras can reach high frame rates, but their dynamic range is typically limited (on the order of $\sim$60~dB). This limitation often causes overexposure in flame regions and suppresses detail in smoke-obscured areas \cite{DEMKO2022112054,PMID:37214717,guan2026}. The resulting information loss can bias particle segmentation and tracking, which reduces the reliability of quantitative size and position estimates. Laser-based diagnostics, such as light-scattering methods, can be precise in optically clear conditions; however, they suffer from signal attenuation and multiple-scattering artifacts in optically dense combustion plumes \cite{Yang2011ParticleSM,JIN2020106066}. Multi-camera systems and complex optical arrangements have also been explored, but they increase hardware complexity and calibration burden, and they can be less tolerant to extreme thermal conditions \cite{1315998,LI2024113342}. Overall, conventional approaches often cannot satisfy extreme HDR, microsecond temporal resolution, and robust imaging through dense smoke within a single practical setup. Consequently, flame regions are frequently overexposed while smoke-obscured areas lose structural detail.
	
	Satisfying these coupled requirements with a single sensor type is impractical. We therefore integrate an SVE camera with a stereo pair of event cameras in a tightly coupled pipeline. Rather than treating the sensors as independent modules \cite{TAO2025}, the pipeline is designed so that each modality supports the other. In this pairing, the SVE branch provides HDR intensity context, whereas the event cameras provide microsecond timing for fast particle dynamics. This intensity context helps interpret event activity in smoke-affected regions, where events alone lack an absolute reference. We then use the SVE-derived HDR output as an intensity prior for event processing, which suppresses smoke-driven artifacts and produces cleaner particle observations for stereo triangulation.
	
	Our core methodological innovations are threefold:
	\begin{itemize}
		\item[$\bullet$] \textbf{Combustion-tailored optical measurement methodology:} We propose a hybrid optical scheme that integrates SVE sensing with a multi-cue smoke-density map to decouple particle radiation from volumetric scattering, yielding calibrated HDR imagery for quantitative tracking.
		\item[$\bullet$] \textbf{Intensity-guided event processing:} We use SVE-derived HDR frames as an intensity reference to filter smoke-induced artifacts in the event stream and classify particle states based on their radiative properties.
		\item[$\bullet$] \textbf{Stereo 3D measurement pipeline:} We construct a stereo pipeline spanning from feature extraction to the estimation of separation height and equivalent radius, verified through geometric calibration and consistency checks across sensing modalities.
	\end{itemize}
	Together, these components permit in-situ, microsecond-resolved observation of boron-based propellant combustion under optically complex conditions.
	The quantitative results are supported by calibration verification and cross-sensor consistency checks, and remaining limitations are discussed in Sec.~4.6.
	
	\begin{figure*}[t]
		\centering
		\includegraphics[width=0.99\textwidth]{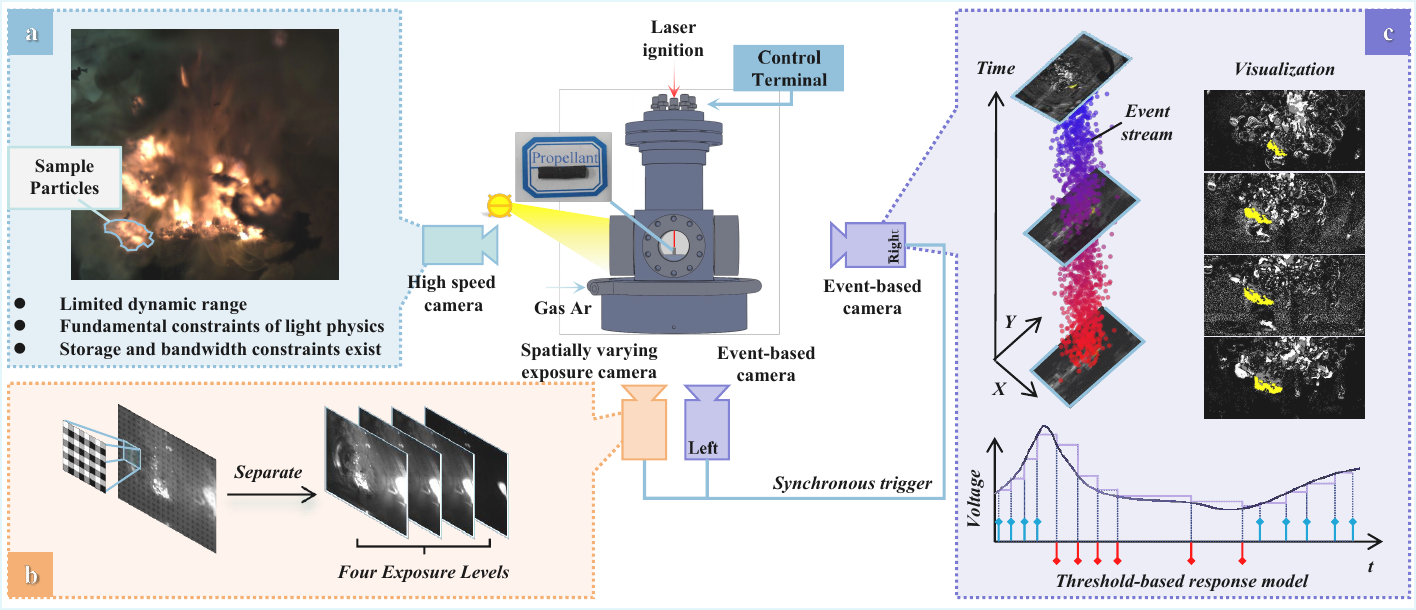}
		\caption{Schematic diagram of optical measurement system.}
		\label{fig:main}
	\end{figure*}
	
	\section{Related Work}\label{sec2}
	Quantitative optical diagnostics for high-energy propellant combustion must address three recurring challenges: intense luminosity, ultrafast dynamics, and dense smoke occlusion. This section surveys recent optical sensing approaches, explains how they handle these challenges in isolation, and summarizes where they fall short in coupled combustion scenes. These limitations motivate the multisensor fusion framework described in Sec.~3.
	
	\subsection{HDR Imaging for Extreme Combustion Environments}\label{sec2.1}
	Combustion diagnostics often involve extreme luminance ranges that can exceed 120~dB. Conventional high-speed cameras can provide adequate frame rates, but their dynamic range is typically limited (about 60--70~dB).
	As a result, bright regions saturate and obscured areas lose detail, which can degrade measurement accuracy.
	
	HDR imaging is therefore widely used to mitigate saturation and recover detail across exposures. Multi-exposure fusion (MEF) \cite{2009Exposure} combines images captured at different exposures, but its sequential capture can introduce motion artifacts and ghosting in fast-changing combustion scenes. Single-shot HDR approaches use specialized sensors to avoid inter-frame motion, including pixel-well-capacity-based \cite{Solid-State}, logarithmic-response \cite{1683901,BOUVIER2014146}, and SVE cameras \cite{SVE,IROS2022}. The SVE camera captures multiple exposures simultaneously via a pixel-level attenuation mask, avoiding motion blur. However, converting the raw mosaic into reliable HDR imagery under dense, scattering smoke is still challenging. While general-purpose MEF algorithms perform well in natural scenes, they are not optimized for the coupled emission and scattering effects prevalent in high-energy propellant combustion. Related fusion studies have addressed scattering-like degradation in other settings, but the combustion case is more tightly coupled with self-emission. Our prior perceptual-fusion work \cite{TAO2025} focused on atmospheric haze and infrared--visible integration, where the volumetric haze is relatively smooth and separable from background structure. In propellant combustion, compact particle emission is intertwined with time-varying volumetric smoke scattering \cite{9746779,Jiang_2023_ICCV}.
	This coupling can cause generic HDR fusion to mix particle radiance with smoke-induced attenuation.
	A combustion-specific strategy is therefore needed to separate particle and smoke cues for downstream measurement \cite{8237767,8839746}. 
	This need motivates the combustion-specific HDR fusion component in our framework.
	
	\subsection{Event-Based Vision for Ultrafast Particle Dynamics}\label{sec2.2}
	Frame-based vision sensors are fundamentally limited by fixed temporal sampling rates, creating a trade-off between motion blur and data throughput. Neuromorphic event cameras operate asynchronously and report logarithmic brightness changes with microsecond temporal resolution and high dynamic range ($>$120~dB) \cite{Survey}. This makes them well suited for high-speed tracking and motion analysis in challenging lighting, with applications in robotics, fluid dynamics, and vibration monitoring \cite{8953722,assessment2025,ams2025,8593561,GUO2025111945,D0LC00556H}.
	
	For quantitative measurement, event cameras also have limitations. A central issue is the lack of absolute intensity: events encode only changes in brightness.
	As a result, appearance-based segmentation and state discrimination are difficult from the event stream alone unless an external intensity reference is available. Reconstructing dense intensity images or repeatable geometric features from sparse, asynchronous events also remains non-trivial \cite{Mueggler2016TheED}. In combustion scenes, turbulent smoke can trigger spurious events that overlap with particle signatures, leading to false detections and tracking errors \cite{10301562}. This ambiguity is addressed by complementary sensing that can provide an intensity reference and scene context for interpreting events.
	
	\subsection{Advanced Optical Metrology in Combustion Diagnostics}\label{sec2.3}
	Optical metrology for combustion diagnostics encompasses diverse approaches, each with distinct advantages and constraints for measuring parameters such as particle size, velocity, and spatial distribution.
	
	High-speed imaging with particle-tracking velocimetry (PTV) provides direct visualization of burning surfaces and particle motion \cite{ZHUO2024119151,LIAO2024113666}, but it still faces difficulty in handling extreme luminosity while freezing ultrafast motion. Laser-based techniques such as DIH \cite{WU20214401}, Mie scattering \cite{CHEN2017225}, PIV, and PLIF \cite{Particle2024} can provide high resolution for specific quantities, but their performance degrades in dense, luminous plumes due to attenuation, multiple scattering, and interference from flame emission. More specialized approaches such as synchrotron X-ray imaging enable penetration through dense media with high spatial resolution \cite{KALMAN2018144}, though facility requirements limit practical application. Ex-situ methods like quench collection allow detailed particle analysis but capture only terminal states rather than dynamic processes \cite{ZHANG202477,GLOTOV202311}.
	
	Collectively, these techniques demonstrate that no single modality satisfies all diagnostic requirements simultaneously.
	Each approach excels in specific aspects, but simultaneous coverage of dynamic range, temporal resolution, and robustness in dense, self-emitting plumes remains difficult.
	This difficulty is amplified in dense smoke and strong self-emission, where visibility and radiometric cues vary rapidly.
	This motivates hybrid sensing that combines complementary strengths and uses calibration-verified geometry for 3D estimation.

	\section{Methodology}\label{sec3}
	Our synergistic metrology framework integrates a tightly coupled sensing setup with a multi-stage processing pipeline. It comprises four core components: (1) optomechanical platform design and calibration; (2) a combustion-optimized multi-exposure HDR fusion algorithm driven by an adaptive, smoke-aware map; (3) event-based particle extraction guided by SVE intensity priors; and (4) multi-view 3D reconstruction and parameter estimation with uncertainty analysis.
	
	\subsection{Integrated Optical Metrology Platform and Calibration}\label{sec3.1}
	To address the challenges posed by extreme dynamic range and ultrafast particle motion in propellant combustion \cite{SVE,Bioinspired}, the platform combines an SVE camera with a stereo pair of neuromorphic event cameras.
	
	The SVE camera (Fig.~\ref{fig:main}(b)) employs a micro-attenuator array mounted in front of the sensor. Each $2\times2$ macro-pixel uses four distinct neutral-density filters, enabling simultaneous capture of four interleaved exposures in a single snapshot. This one-shot mechanism avoids motion artifacts inherent to multi-frame HDR techniques. The event cameras (Fig.~\ref{fig:main}(c)) operate asynchronously: each pixel detects logarithmic intensity changes and outputs events $e(u,v,p,t)$ with microsecond temporal resolution and dynamic range exceeding 120~dB \cite{Survey,global}, where $(u,v)$ are pixel coordinates, $p\in\{+1,-1\}$ denotes polarity, and $t$ is the timestamp.
	
	A critical prerequisite for cross-modal fusion and accurate 3D metrology is precise spatiotemporal calibration.
	
	\emph{1) Intrinsic and Extrinsic Calibration:}
	Intrinsic parameters (focal length, principal point, distortion) of the SVE and event cameras are individually estimated using Zhang's \cite{Zhang} method with a planar checkerboard.
	For the stereo configuration, the world frame is defined as the left event-camera frame (i.e., $\mathbf{R}^l=\mathbf{I}$ and $\mathbf{t}^l=\mathbf{0}$).
	With the world-to-camera model $\mathbf{X}^c=\mathbf{R}^c\mathbf{X}^w+\mathbf{t}^c$, the right-camera extrinsics $(\mathbf{R}^r,\mathbf{t}^r)$ denote the rigid transform from the left (world) frame to the right camera frame.
	A shared calibration target visible to all sensors is used to estimate the rigid transformation between the SVE and stereo-event systems, enabling spatial registration of SVE-derived intensity priors onto the event views.
	
	\emph{2) Hardware-Level Synchronization:}
	A hardware trigger synchronizes the SVE exposure and both event cameras. The trigger also injects a special timestamped event into each event stream, providing a global time reference ($t=0$) for the corresponding SVE frame.
	Synchronization precision was verified to be within $12~\mu$s. For the particle velocities observed in our experiments, this residual offset corresponds to sub-pixel displacement, supporting accurate cross-modal alignment.
	
	\emph{3) SVE Image Reconstruction:} The camera's raw output is a single mosaic image containing the four spatially interleaved exposures. This image is first separated into four sub-images via spatial demultiplexing. To compensate for reduced spatial sampling and ensure alignment, we employ a reconstruction process similar to color demosaicing: a gradient-corrected interpolation algorithm is applied to each sub-image, yielding a set of full-resolution, spatially registered images ready for fusion and analysis \cite{1326587,IROS2022}.
	
	\subsection{Combustion-Specific HDR Image Formation}\label{sec3.2}
	Propellant combustion involves intense point-source radiation from particles and dense, spatially varying smoke. Conventional MEF methods often struggle under these conditions.
	
	\emph{1) Smoke-likelihood estimation for SVE exposures.} 
	Given the quad-exposure sequence $\{I_k\}_{k=1}^K$, the model constructs a smoke-likelihood indicator $F(x)$ from four complementary features.
	
	To characterize radiance variations across exposures while suppressing noise in dark regions, we first define the brightness deviation intensity
	\begin{equation}
		BI(x) = \frac{1}{K}\sqrt{\sum_{k=1}^K (\max(I_k(x),T) - \mu_k)^2},\quad T = \mu_k/2,
		\label{eq:BI}
	\end{equation}
	where $I_k(x)$ is the pixel intensity at location $x$ in the $k$-th exposure, and $\mu_k$ is the mean intensity of that exposure. The threshold $T$ limits the influence of low-SNR pixels, stabilizing the descriptor in strongly underexposed regions.
	
	To retain particle edges and fine textures under varying illumination, we adopt the average Weber contrast
	\begin{equation}
		WC(x) = \frac{1}{K}\sum_{k=1}^K \frac{|\nabla I_k(x)|}{I_k(x) + 1},
		\label{eq:WC}
	\end{equation}
	where $\nabla$ denotes the gradient operator. This formulation emphasizes local contrast relative to background brightness and is particularly effective for highlighting burning-particle boundaries embedded in low-contrast smoke.
	
	Non-uniform smoke distribution and combustion glare are modeled using the atmospheric scattering formulation
	\begin{equation}
		I(x) = J(x)t(x) + A(1 - t(x)),
		\label{eq:Model}
	\end{equation}
	with smoke-free radiance $J(x)$, transmission $t(x)$, and atmospheric light $A$. Assuming locally constant $t(x)$ and $A$, we define dark and bright channels over the multi-exposure stack \cite{2017ICIVC,DCP}:
	\begin{equation}
		\left\{\begin{array}{l}
			I^{\mathrm{d}}(x) = t \cdot J^{\mathrm{d}}(x) + A(1 - t)\\
			I^{\mathrm{b}}(x) = t \cdot J^{\mathrm{b}}(x) + A(1 - t)
		\end{array}\right.
		\label{eq:Jd_Jb}
	\end{equation}
	where $I^{\mathrm{d}}(x)$ and $I^{\mathrm{b}}(x)$ are obtained by applying minimum/maximum operators across the four exposures. A generalized contrast feature is then defined as
	\begin{equation}
		CF(x) = 1 - \frac{I^{\mathrm{d}}(x)}{\max(I^{\mathrm{b}}(x),1)},
		\label{eq:CF}
	\end{equation}
	so that large $CF(x)$ values indicate regions where the dark channel is strongly suppressed relative to the bright channel, which is more consistent with optically thick smoke than with compact, emissive particles.
	
	\begin{figure*}[t]
		\centering
		\includegraphics[width=0.66\textwidth]{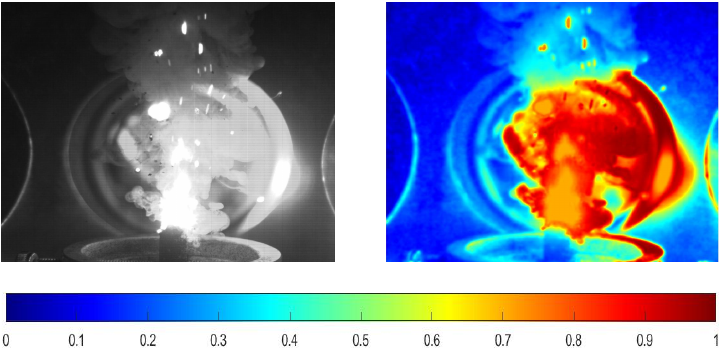}
		\caption{Probabilistic smoke-likelihood map $F(x)$, with the color bar indicating a normalized visibility score (higher implies stronger smoke influence and reduced visibility).}
		\label{fig:smoke}
	\end{figure*}
	
	To detect regions with strong inter-exposure fluctuations, typical of flame cores and specular highlights, we compute the normalized pixel response variance
	\begin{equation} 
		V(x) = \frac{\mathrm{Var}(x) - \bar{\chi}}{\bar{\chi} + \varepsilon},
		\label{eq:V}
	\end{equation}
	where $\mathrm{Var}(x)$ is the pixel variance across different exposures, $\bar{\chi}$ is the geometric mean variance of the image, and $\varepsilon$ is a small constant ensuring numerical stability. This feature serves as a robust cue for HDR-saturated or highly non-linear regions that require cautious fusion. 
	
	The dimensionless smoke-likelihood indicator $F(x)$ is synthesized from a weighted combination of these four features:
	\begin{equation}
		F(x) = \alpha\,BI(x) + \beta\,WC(x) + \gamma\,CF(x) + \sigma\,V(x),
		\label{eq:model}
	\end{equation}
	with $\alpha+\beta+\gamma+\sigma=1$. In our combustion experiments, the coefficients are empirically set to $(\alpha,\beta,\gamma,\sigma)=(0.1,0.4,0.2,0.3)$, with higher weights on edge preservation and saturation handling (larger $\beta,\sigma$) and lower weights on absolute brightness terms (smaller $\alpha,\gamma$). The resulting probabilistic smoke-likelihood map $F(x)$, illustrated in Fig.~\ref{fig:smoke}, is then segmented into $M=4$ regions ${P_m}_{m=1}^{M}$ via histogram-based thresholding followed by Gaussian mixture model fitting \cite{2019Scene}. These regions approximately correspond to dense smoke, transition zones, and areas dominated by flames and particles, forming the basis for subsequent sub-regional fusion.
	
	\emph{2) Region-wise Retinex fusion guided by ${\rm F}(x)$.}
	With the scene segmented by $F(x)$, each exposure $I(x)$ is decomposed into an illumination $L(x)$ and a reflection $R(x)$ layer using a Retinex-type model \cite{WGIF,Simultaneous}:
	\begin{equation}
		I(x) = L(x)\cdot R(x).
	\end{equation}
	This separation allows for smoke suppression in the low-frequency illumination layer and particle detail enhancement in the high-frequency reflection layer.
	
	Region-specific fusion weights for the illumination layer are calculated using a dual-weighting mechanism:
	\begin{equation}
		\label{eq:w1}
		W_{L_1,k}(x) = 
		\frac{\mathrm{Grad}_k(L_k(x))^{-1}}
		{\sum_{k=1}^{K} \mathrm{Grad}_k(L_k(x))^{-1} + \varepsilon},
	\end{equation}
	\begin{equation}
		\label{eq:w2}
		W_{L_2,k}(x) = 
		\frac{1}{\psi}\exp\!\left(-\frac{(L_k(x)-u_k)^2}{2\delta^2}\right),
	\end{equation}
	where $\mathrm{Grad}_k(L_k(x))$ is the gradient of the cumulative histogram of $L_k(x)$, serving as a surrogate for exposure quality; $u_k$ is the regional mean brightness in exposure $k$, $\psi$ is a normalization coefficient, $\delta$ controls weight decay, and $\varepsilon$ is a small constant. Final illumination weights are:
	\begin{equation}
		W_{L,k}(x) = W_{L_1,k}(x)\cdot W_{L_2,k}(x).
	\end{equation}
	
	\begin{figure*}[t]
		\centering
		\includegraphics[width=0.9\textwidth]{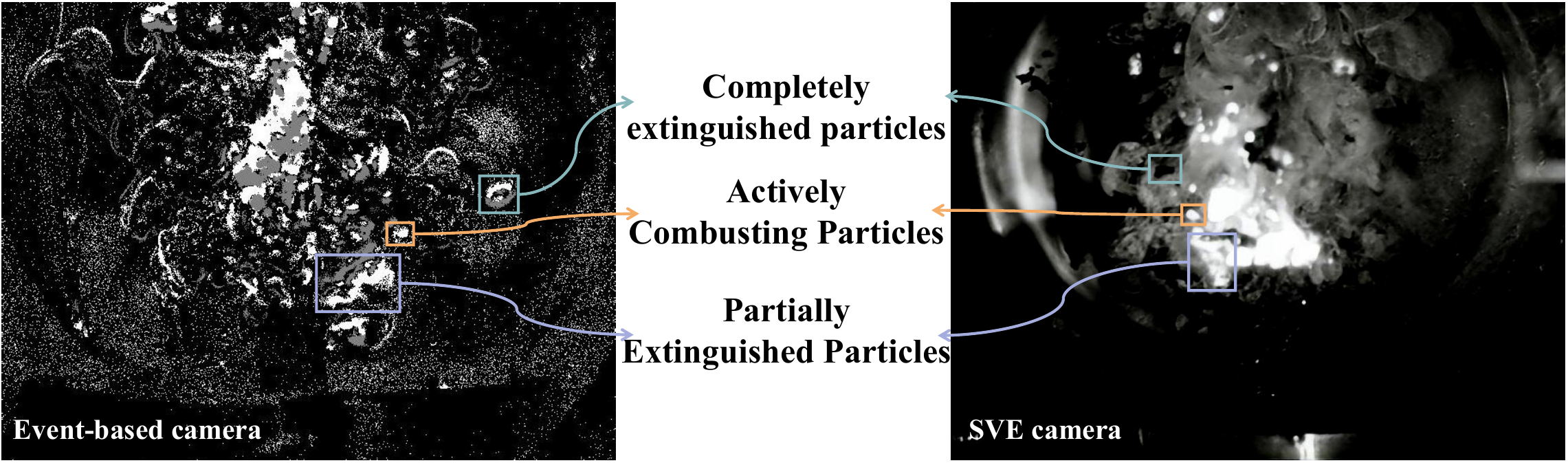}
		\caption{Particle Imaging Extraction Strategy Diagram.}
		\label{fig:particle}
	\end{figure*}
	
	Fusion is performed at each pyramid level $l$:
	\begin{equation}
		\label{eq:pyr_fusion}
		\hat{I}^{(l)} = 
		\bigl\{\mathcal{L}^{(l)}\{{\rm L}_k\}\cdot\mathcal{G}^{(l)}\{{\rm W_{L},k}\}\bigr\}
		\bigl\{\mathcal{L}^{(l)}\{{\rm R}_k\}\cdot\mathcal{G}^{(l)}\{{\rm W_{R},k}\}\bigr\},
	\end{equation}
	Let $\hat{I}^{(l)}(x)$ denote the fused image at pyramid level $l$. $\mathcal{G}^{(l)}(\cdot)$ and $\mathcal{L}^{(l)}(\cdot)$ are Gaussian and Laplacian pyramid operators. The reconstructed HDR image provides a high-contrast, smoke-suppressed representation of the combustion scene.
	
	As summarized in Fig.~\ref{fig:particle}, this HDR output serves as a robust intensity prior for subsequent event-based particle extraction, enabling accurate tracking and metrology even in heavily smoke-obscured environments. The performance of the combustion-specific HDR formation, particularly its capability to retain critical radiometric and structural features for downstream kinematic analysis, is quantitatively evaluated in Section~4.
	
	\subsection{Event-Based Particle Extraction with SVE Guidance}\label{sec3.3}
	To capture transient particle evolution under smoke-obscured HDR conditions, an event-based extraction scheme is developed using the HDR prior from Section~\ref{sec3.2}. While event cameras offer microsecond resolution and $>$120~dB dynamic range, their differential response is susceptible to smoke-induced artifacts. The HDR image adds spatial context to disambiguate events and identify particle structures.
	
	Particles are categorized into three classes based on radiative and kinematic states:
	
	\emph{(1) Actively combusting particles.}
	These particles exhibit strong self-emission and rapid motion, generating dense positive-polarity events at the core and negative-polarity events along the motion trail. For geometry extraction, it is sufficient to retain the positive-polarity subset $\{ ({u_i},{v_i})\mid {p_i} =  + 1\} $, which closely approximates the luminous particle body. Spatial consistency with bright regions in the HDR image is enforced to reduce spurious events.
	
	\emph{(2) Completely extinguished particles.}
	Once combustion ceases, residual particles generate only weak events, primarily at the particle--background interface due to contrast changes. Both polarities are collected within a short time window, and a polygonal envelope $\Omega$ is delineated to enclose the mixed-polarity boundary events. The set $\{(u_i,v_i)\mid (u_i,v_i)\in\Omega\}$ defines the 2D support of the extinguished particle and is further validated against the corresponding low-intensity, high-contrast region in the HDR image.
	
	\begin{figure*}[t]   
		\centering
		\includegraphics[width=0.75\textwidth]{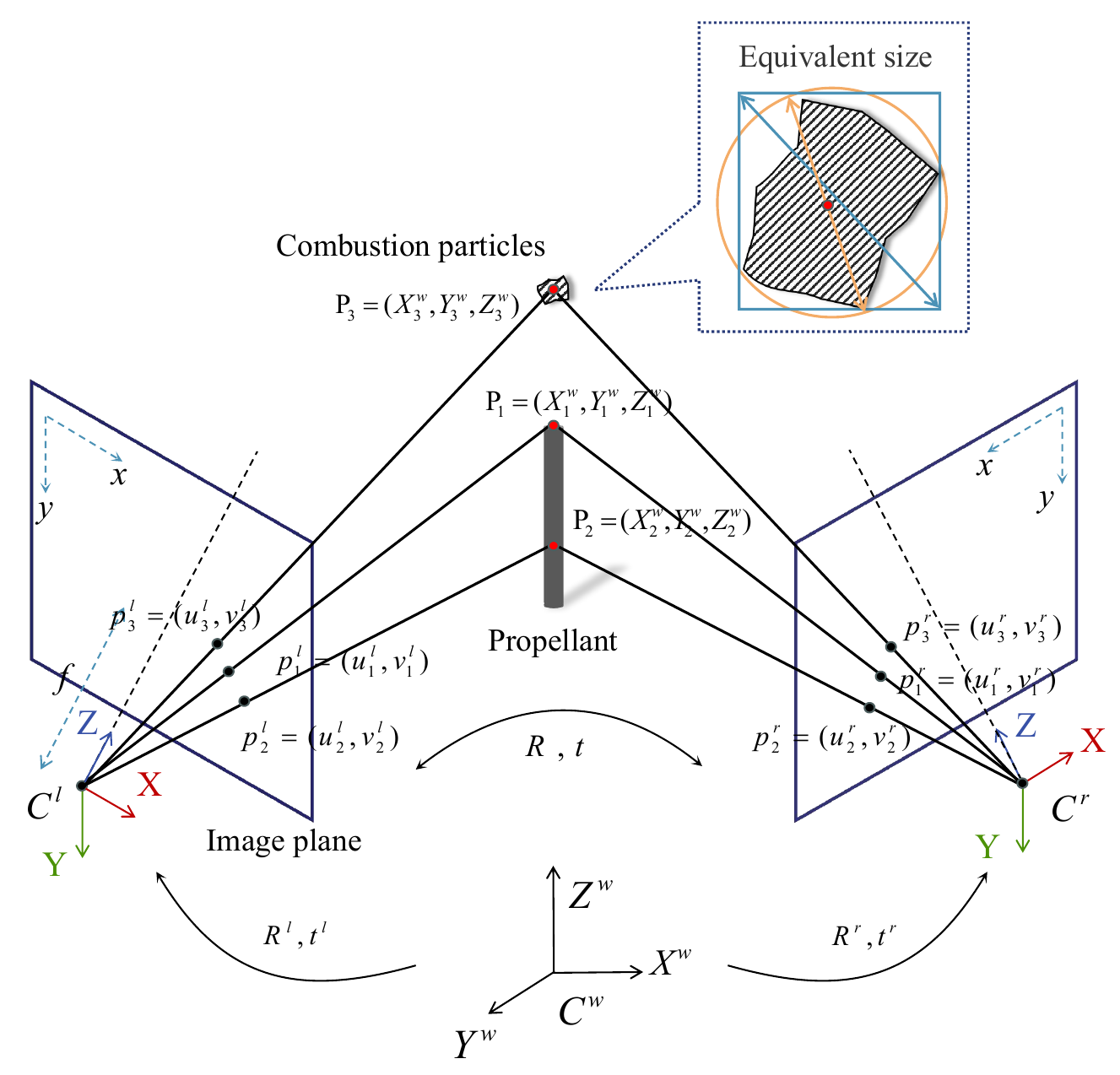}
		\caption{Propellant combustion stereo imaging model.}
		\label{fig:2D}
	\end{figure*}  
	
	\emph{(3) Partially extinguished particles.} 
	During extinction, residual combustion and thermal gradients can produce hybrid-polarity events distributed across both edges and interiors. We estimate instantaneous velocity by tracking the tail contour within a short temporal window and compensate for motion by shifting events according to a local displacement field $\Phi$, implemented as a locally constant-velocity warp estimated per event view. For events in $[t_0,t_0+\Delta t]$, the tail-contour displacement $\Delta\mathbf{p}$ is measured between two equal sub-windows, and $\mathbf{v}=\Delta\mathbf{p}/\Delta t$ is computed.
	The warp is defined as $\Phi(\mathbf{p}_i,t_i)=\mathbf{v}(t_i-t_0)$, and event coordinates are compensated as $\mathbf{p}'_i=\mathbf{p}_i-\Phi(\mathbf{p}_i,t_i)$, where $\mathbf{p}_i=[u_i,v_i]^{\mathrm{T}}$. The compensated set $\{(\mathbf{p}'_i,p_i,t_i)\}$ generates a stabilized particle footprint, which helps reconstruct complete particle trajectories despite fast motion and intermittent visibility.
	
	Smoke interference remains a major challenge: turbulent, semi-transparent smoke can trigger particle-like event textures \cite{Numerical,BELAL2018410}, leading to contour aliasing and reduced discriminability. To mitigate this, we apply SVE-based HDR guidance as spatial gating: only event clusters that (i) lie in regions with sufficient visibility according to $F(x)$ and (ii) show particle-like radiative signatures are retained as candidates. Here, $F(x)$ is used for visibility-based gating rather than as a physically calibrated smoke measurement. The contour-based segmentation threshold primarily governs boundary tightness, whereas candidate acceptance is dictated by HDR-supported gating and stereo consistency. This filtering reduces smoke-driven events and provides clean inputs for the stereo reconstruction and metrology pipeline in Section~\ref{sec3.4}.
	
	\subsection{Multi-View Particle Metrology}\label{sec3.4}
	To quantitatively characterize the 3D evolution of combustion particles, a stereo event-based metrology system is constructed (Fig.~\ref{fig:2D}). Two synchronized event cameras observe the propellant from different viewpoints, and the SVE-guided extraction in Section~\ref{sec3.3} provides clean 2D particle contours in each view. This multi-view configuration enables precise measurement of separation height and equivalent particle size with microsecond temporal resolution, offering critical data for analyzing and optimizing propellant combustion \cite{10497132,Modeling,Unsupervised}.
	
	\emph{1) 3D reconstruction and separation height.}
	For a particle centroid observed in both views, its 3D position is obtained by triangulation.
	Let $\mathbf{P}^w=[X^w,Y^w,Z^w,1]^{\mathrm{T}}$ denote a point in world coordinates, where the world frame is defined as the left event-camera frame (i.e., $\mathbf{R}^l=\mathbf{I}$ and $\mathbf{t}^l=\mathbf{0}$ by definition).
	Let $\mathbf{P}^l=[u^l,v^l,1]^{\mathrm{T}},\mathbf{P}^r=[u^r,v^r,1]^{\mathrm{T}}$ denote its projections on the left and right image planes.
	We adopt the convention that the extrinsics map a 3D point from the world frame to the camera frame, i.e., $\mathbf{X}^c=\mathbf{R}^c\mathbf{X}^w+\mathbf{t}^c$.
	The projection from world to the right camera is
	\begin{equation}
		{s^r}\left[ {\begin{array}{c}
				{{u^r}}\\
				{v^r}\\
				1
		\end{array}} \right]
		=
		{{\bf{K}}^r}\left[ {{{\bf{R}}^r}\,|\,{{\bf{t}}^r}} \right]
		\left[ {\begin{array}{c}
				{{X^w}}\\
				{{Y^w}}\\
				{{Z^w}}\\
				1
		\end{array}} \right],
	\end{equation}
	and analogously for the left camera.
	Here, $s^l$ and $s^r$ are depth scaling factors, $\mathbf{K}^{l,r}$ are intrinsic matrices, and $(\mathbf{R}^{l,r},\mathbf{t}^{l,r})$ are extrinsic parameters obtained from stereo calibration, with $(\mathbf{R}^r,\mathbf{t}^r)$ representing the rigid transform from the world frame (left camera) to the right camera frame.
	
	Accurate 3D coordinates of particle centroids are essential for trajectory reconstruction. To define a physically meaningful separation height, the known vertical orientation of the propellant specimen is exploited. As shown in Fig.~\ref{fig:2D}, points $\mathbf{P}_1$ and $\mathbf{P}_2$ on the combustion column define the axial direction, and $\mathbf{P}_3$ denotes the reconstructed particle centroid. The column axis unit vector is $\hat{\mathbf{n}}=(\mathbf{P}_2-\mathbf{P}_1)/\|\mathbf{P}_2-\mathbf{P}_1\|$, and the separation height $\Delta h$ is defined as the axial projection of $\mathbf{P}_3-\mathbf{P}_1$:
	\begin{equation}
		\Delta h = {\left( {{{\bf{P}}_3} - {{\bf{P}}_1}} \right)^{\rm T}}\widehat {\bf{n}}.
	\end{equation}
	This formulation directly links the 3D reconstruction to the combustion geometry and enables consistent comparison of particle lift-off behavior across tests.
	
	\begin{figure*}[tp]
		\centering
		\includegraphics[width=0.7\textwidth]{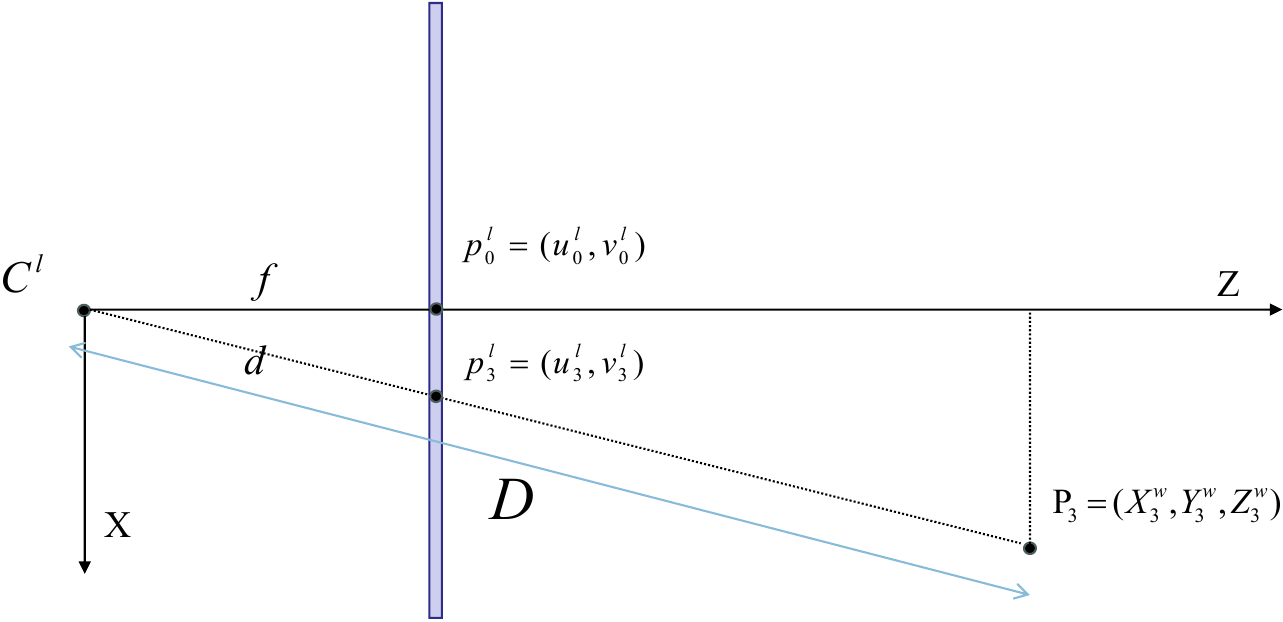}
		\caption{Physical Scale Calculation Model.}
		\label{fig:SCALE}
	\end{figure*}
	
	\emph{2) Physical scaling and equivalent particle size.}
	To characterize size-dependent dynamics, the \emph{equivalent radius} $r_e$ is used as a scalar descriptor. This metric is defined as the radius of a circle whose area matches the measured particle projection. Given that boron agglomerates are irregular, their apparent contour varies with viewpoint and focus. Consequently, we report a contour-based equivalent radius derived from stereo-confirmed particles, serving as a statistical descriptor rather than a view-invariant geometric truth.
	
	To convert pixel measurements into physical units, we employ a pinhole-based spatial scale model (Fig.~\ref{fig:SCALE}). The scale factor $S_f$ (mm/pixel) relates image-plane distances to real-world lengths:
	\begin{equation}
		S_f = \frac{D\cdot\mu}{d},
	\end{equation}
	where $D$ is the distance from the object to the camera optical center, $\mu$ is the pixel pitch, and $d$ is the object distance along the optical axis, computed as
	\begin{equation}
		d = \sqrt{f^2 + \|\mathbf{p}_3^l-\mathbf{p}_0^l\|^2},
	\end{equation}
	with $f$ the focal length in pixels and $\mathbf{p}_3^l$, $\mathbf{p}_0^l$ the image coordinates of the particle and principal point in the left view, respectively. This depth-aware scaling avoids the bias that would arise from assuming a constant scale at all depths.
	
	For each stereo-matched particle, the physical cross-sectional area $S$ is computed by integrating the event-based contour region and applying the local scale factor. Instead of using bounding boxes, which tend to overestimate the size of elongated fragments \cite{2022Experimental}, we derive the equivalent radius $r_e$ directly from the area:
	\begin{equation}
		r_e = \sqrt{\frac{S}{\pi}},
	\end{equation}
	where $S$ is the physical area in mm$^2$. When an equivalent diameter is reported, it is defined as $d_e=2r_e$. Although real particles are not perfectly spherical, this circular-equivalent metric provides a consistent and reproducible scalar descriptor suitable for statistical analysis of particle size distributions.
	
	\section{Experiments and Analysis}\label{sec4}
	Experiments were carried out on boron-based propellant combustion to assess the proposed SVE--event metrology framework. The evaluation covers five complementary aspects: HDR fusion quality in the SVE branch, preservation of metrology-relevant features, event-based verification of separation-height measurement, cross-modal consistency of equivalent-size estimates, and statistical characterization of the size distribution.
	
	\begin{figure*}[bp]
		\centering
		\includegraphics[scale=0.24]{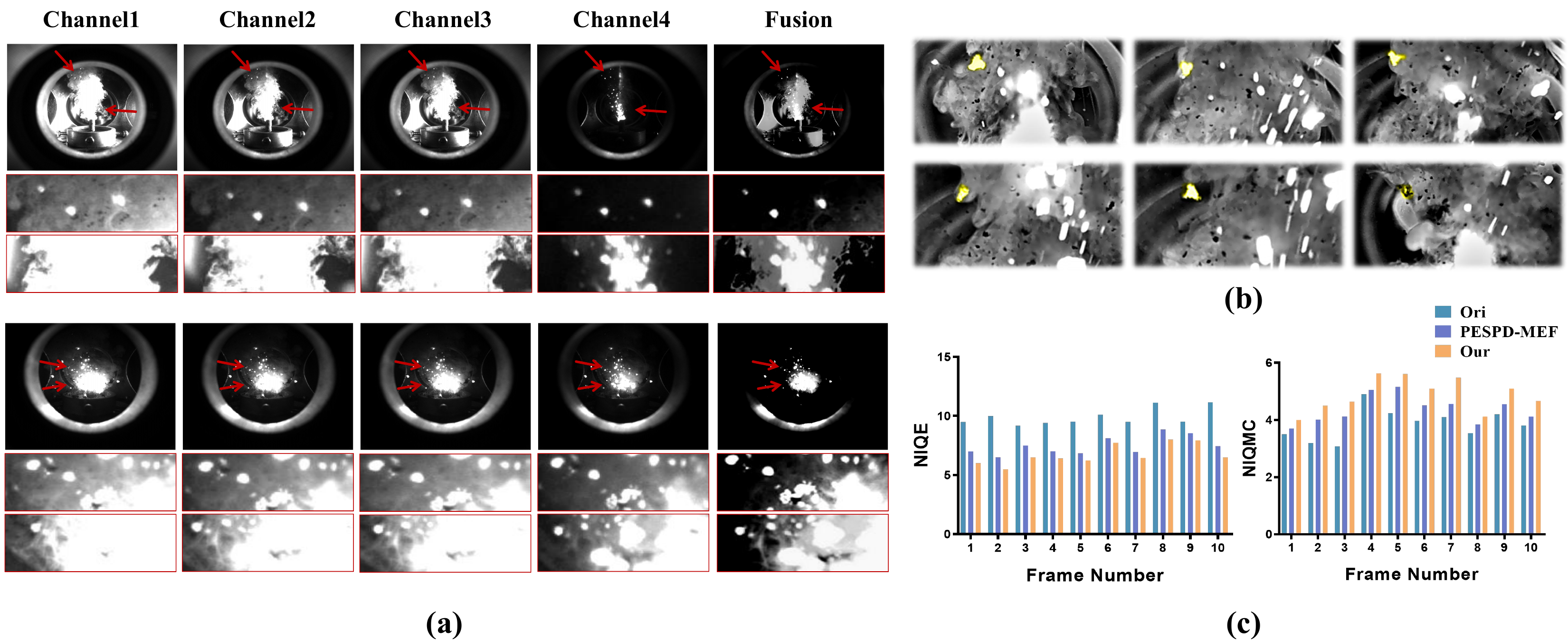}
		\caption{SVE HDR display: (a) Comparison of results before and after fusion. (b) Multi-state tracking of specific particles. (c) Comparison of indicators before and after fusion. }
		\label{fig:data1}
	\end{figure*}
	
	\subsection{Evaluation of Smoke-Aware HDR Fusion}\label{sec4.1}
	Fig.~\ref{fig:data1} illustrates the behavior of the smoke-aware HDR fusion pipeline on representative combustion frames. The raw SVE channels exhibit substantial exposure imbalance: low-transmittance channels retain the bright plume core but suppress background details, whereas high-transmittance channels preserve the launch structure and weak smoke patterns but saturate in the brightest regions. The fused result in Fig.~\ref{fig:data1}(a) combines these complementary cues, keeping particle morphology while still showing the surrounding smoke-related structures.
	
	Across the ignition transient and the quasi-steady phase, the fused SVE sequence maintains usable contrast and spatial detail.
	This supports continuous tracking of individual particles in Fig.~\ref{fig:data1}(b). Particle silhouettes remain separable even near steep brightness transitions, which is important for stable contour extraction and subsequent multi-view reconstruction.
	
	For a no-reference quality check, Fig.~\ref{fig:data1}(c) reports NIQE~\cite{niqe} and NIQMC~\cite{niqmc} for raw SVE inputs, PESPD-MEF~\cite{PESPD-MEF}, and our method.
	Across the sampled frames, our fusion yielded lower NIQE and higher NIQMC than both the raw SVE inputs and PESPD-MEF. These metrics reflect perceived image quality rather than metrology accuracy, but they align with the visual reduction of saturation and noise in the fused HDR output. In the overall pipeline, the fused HDR output serves as an intensity reference for event-based particle segmentation (Fig.~\ref{fig:particle}), which helps stabilize the inputs to stereo measurement.
	
	\begin{figure*}[tp]
		\centering
		\includegraphics[width=0.8\textwidth]{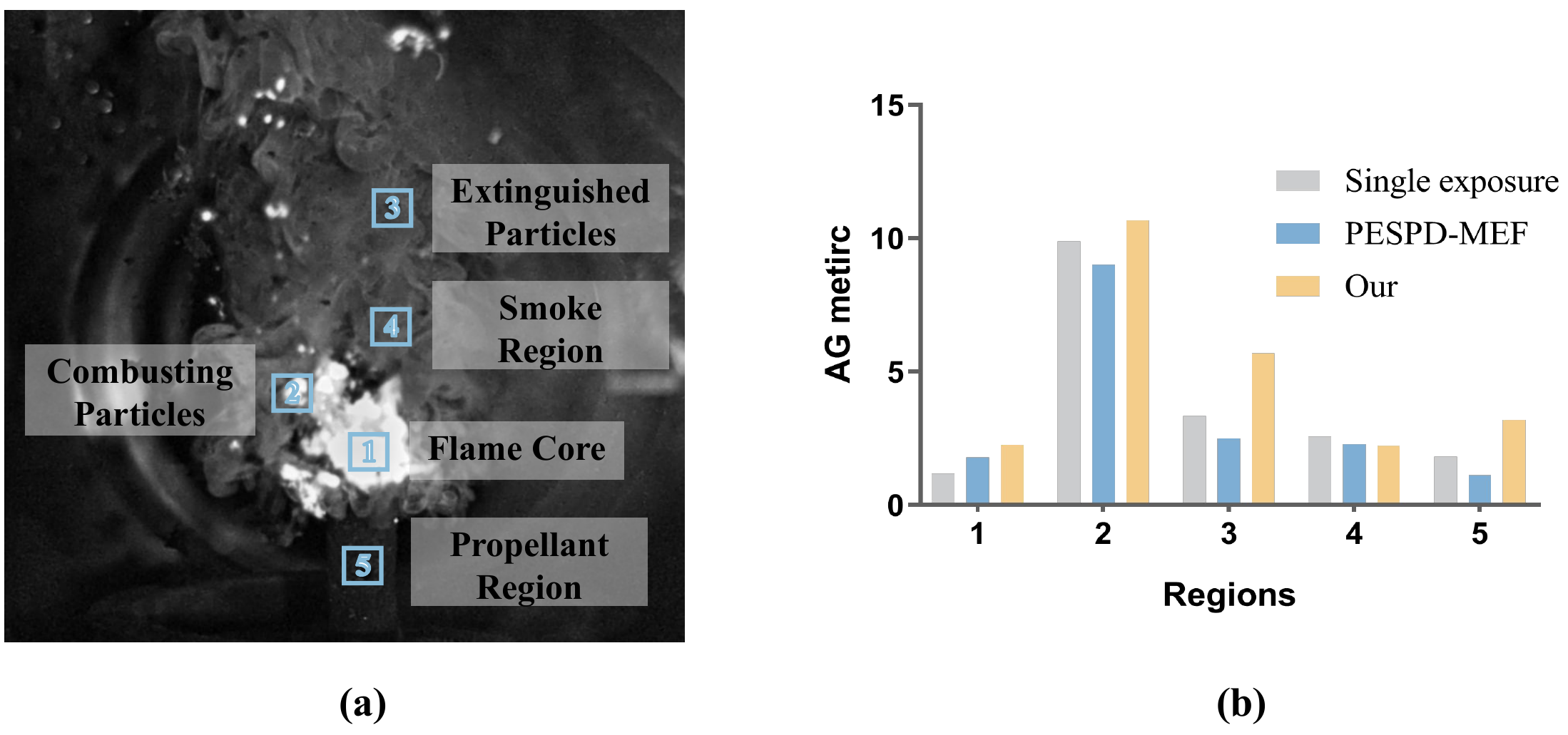}
		\caption{Quantitative feature preservation evaluation. (a) Combustion scene with five Regions of Interest (ROIs): 1) Flame Core, 2) Combusting Particles, 3) Extinguished Particles, 4) Smoke Region, 5) Propellant Region. (b) Average Gradient (AG) comparison of different methods across these ROIs.}
		\label{fig:ag_comparison}
	\end{figure*}
	\begin{table}[tp]
		\small
		\centering
		\renewcommand{\arraystretch}{1.25}
		\caption{Average Gradient (AG) values for different methods across selected ROIs. Bold values indicate the optimal result.}
		\begin{tabular}{cccc}
			\toprule
			Region of Interest & Best Single Exp. & PESPD-MEF~\cite{PESPD-MEF} & \textbf{Our Method} \\
			\midrule
			ROI 1 (Flame Core) & 1.180 & 1.787 & \textbf{2.250} \\
			ROI 2 (Combusting Particles) & 9.886 & 9.009 & \textbf{10.664} \\
			ROI 3 (Extinguished Particles) & 3.346 & 2.505 & \textbf{5.690} \\
			ROI 4 (Smoke Region) & \textbf{2.562} & 2.279 & 2.216 \\
			ROI 5 (Propellant Region) & 1.811 & 1.116 & \textbf{3.187} \\
			\bottomrule
		\end{tabular}
		\label{tab:ag_results}
	\end{table}
	
	\subsection{Feature Preservation for Quantitative Metrology}\label{sec4.2}
	While full-frame image quality metrics (e.g., NIQE and NIQMC in Fig.~\ref{fig:data1}(c)) provide a general assessment, a task-specific evaluation is essential to verify the preservation of features critical for quantitative combustion diagnostics. Reliable tracking and contour fitting depend on clear edges and sufficient local contrast in regions containing particles. To quantify this aspect, average gradient (AG) was computed on five regions of interest (ROIs) in Fig.~\ref{fig:ag_comparison}(a) and compared across methods. These ROIs cover representative scene components: (1) flame core, (2) combusting particle agglomerates, (3) extinguished particles, (4) optically thick smoke, and (5) the propellant column. AG, used here as a simple indicator of local sharpness, was measured for the best single exposure, PESPD-MEF~\cite{PESPD-MEF}, and our method.
	
	\begin{figure*}[tp]   
		\centering
		\includegraphics[width=0.9\textwidth]{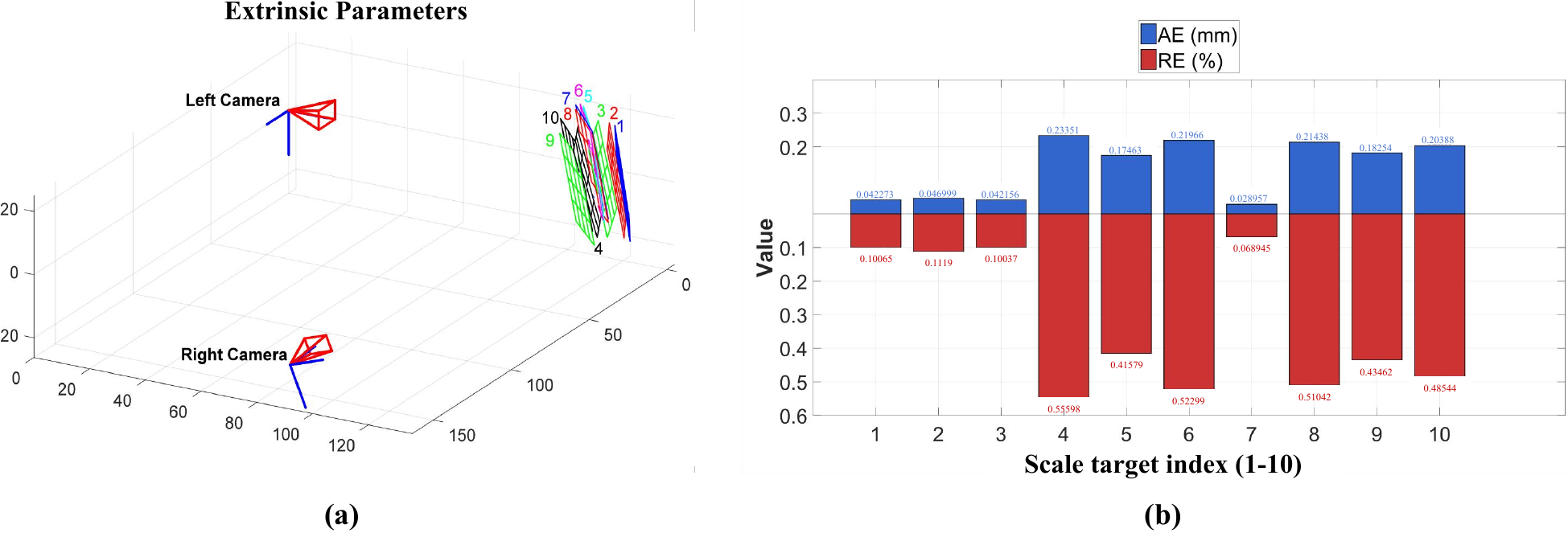}
		\caption{Overview of stereo event-camera calibration and validation: (a) Extrinsic parameters of the stereo event-camera observation. (b) Stereo triangulation accuracy validation using reference scale targets. Indices 1--10 denote ten independent calibration scale targets with known physical sizes used for accuracy validation.}
		\label{fig:cameras}
	\end{figure*} 
	\begin{table*}[tp]
		\centering
		\renewcommand{\arraystretch}{1.25}
		\small
		\caption{Calibration results of our stereo event-camera observation system.}
		\begin{tabular}{c c c}
			\toprule
			Event Cameras & Left Camera & Right Camera\\
			\midrule
			Focal Length (pixels) & $1703.25$ & $1959.47$\\
			Principal Point (pixels) & $(582.69, 497.77)$ & $(502.29, 387.05)$\\
			Distortion Parameter & $(-0.11, 0.22, -0.01,
			0.01, 0.00)$ & $(0.13, -5.68, -0.02,
			0.01, 0.00)$\\
			Euler Angle $(\mathrm{rad})$ & $(0, 0, 0)$ & $(-0.360, 1.410, 0.171)$\\
			Translation Vector $(\mathrm{mm})$ & $(0, 0, 0)$ & $(-105.90, -28.25, 149.75)$\\
			\bottomrule
		\end{tabular}
		\label{tab:stereo-event}
	\end{table*}
	\begin{table*}[tp]
		\centering
		\small
		\renewcommand{\arraystretch}{1.25}
		\caption{Measurement error of our stereo-event camera observation system.}
		\scalebox{0.95}{
			\begin{tabular}{c c c c}
				\toprule
				AE Mean (mm) & AE STD (mm) & RE Mean (\%)  & RE STD (\%)  \\ $0.0659$ & $0.1554$ & $0.3307$ & $0.2067$\\
				\bottomrule
		\end{tabular}}
		\label{tab:Measurement}
	\end{table*}
	
	Fig.~\ref{fig:ag_comparison}(b) and Table~\ref{tab:ag_results} show that our fusion increases AG in most ROIs with particle- or structure-related edges. For example, in actively combusting particle agglomerates (ROI~2), our method attained an AG of 10.664, outperforming PESPD-MEF (9.009) by 18.4\% and the best single exposure (9.886) by 7.9\%. Similarly, in the intense flame core (ROI~1) and propellant region (ROI~5), our method substantially enhances feature preservation. Notably, while superior AG is observed in particle-dominated regions, the optically thick smoke region (ROI~4) shows a moderately lower AG (2.216) than the best single exposure (2.562). This behavior reflects the design intention of our combustion-specific HDR fusion, which actively suppresses diffuse smoke interference and decouples smoke signatures from particle radiation. Reducing gradients from low-contrast smoke can make particle boundaries more distinguishable in the subsequent pipeline. Coupled with HDR-guided gating, this suppresses smoke-induced spurious events and bolsters metrological stability. These results confirm that the fusion step not only extends the usable dynamic range but also preserves the edge sharpness on which contour-based segmentation and tracking depend.
	
	\begin{figure*}[tp]
		\centering
		\includegraphics[width=0.93\textwidth]{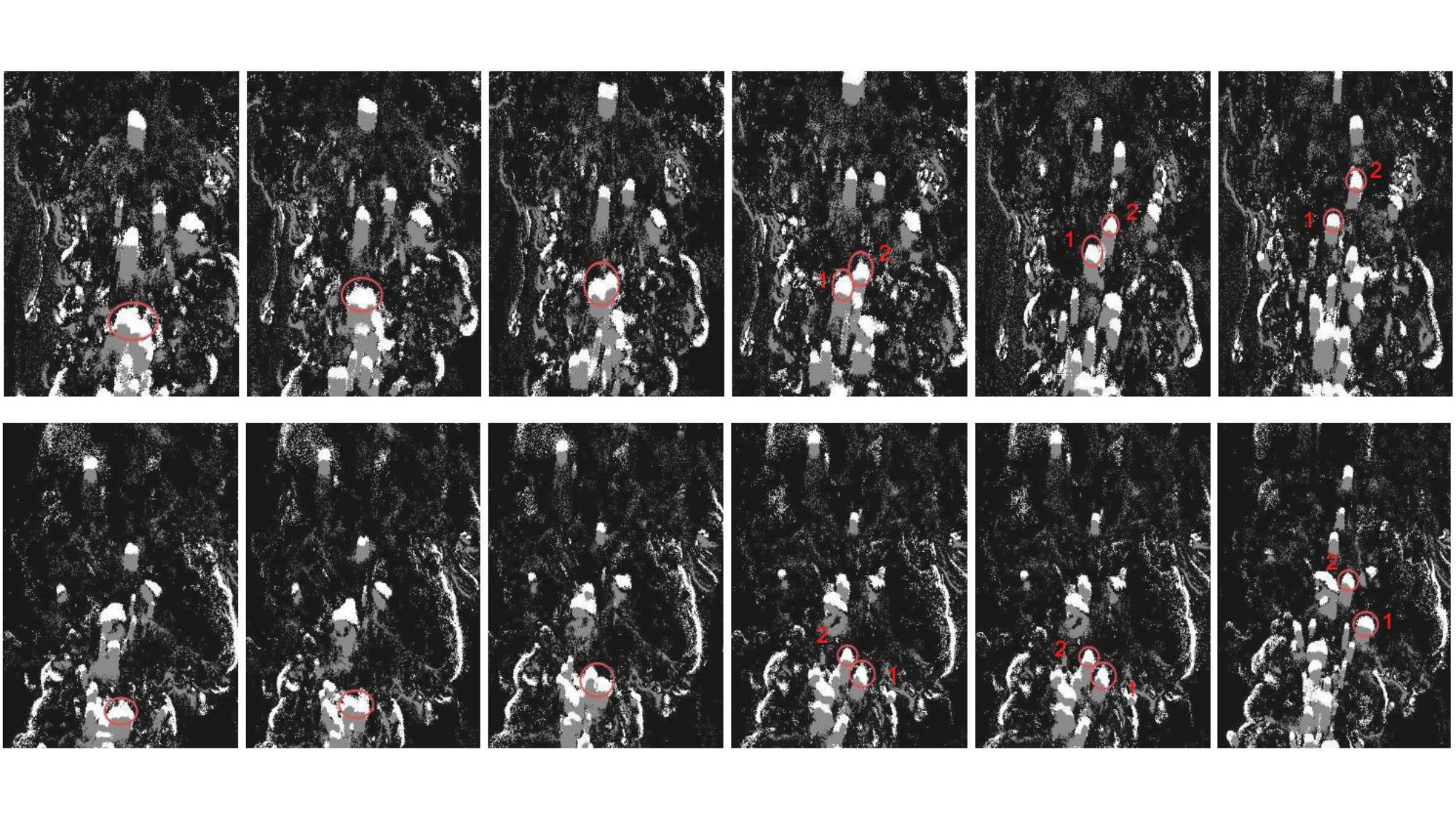}
		\caption{Stereo event-camera observation of particle monitoring.}
		\label{fig:particle-event}
	\end{figure*}
	
	\begin{figure*}[tp]
		\centering
		\includegraphics[width=0.8\textwidth]{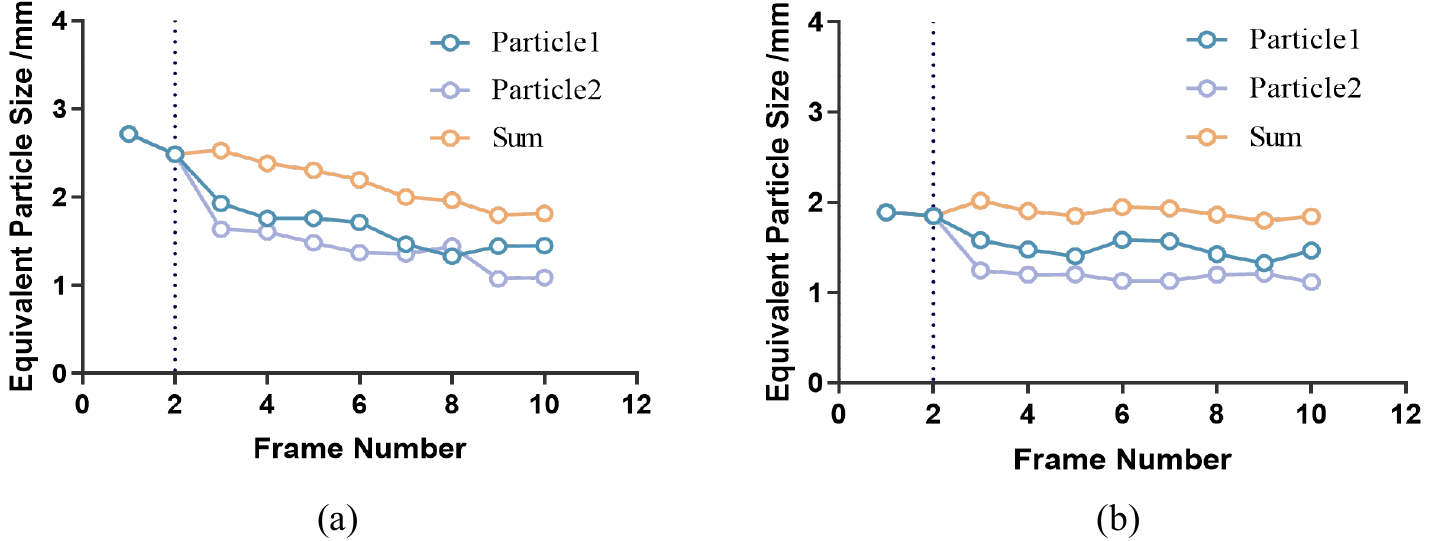}
		\caption{Particle separation monitoring: (a) measurement results from the left event camera; (b) measurement results from the right event camera. The dotted line indicates the particle separation time point.}
		\label{fig:data2}
	\end{figure*}
	
	\subsection{Event-Based Separation Measurement}\label{sec4.3}
	Table~\ref{tab:stereo-event} and Fig.~\ref{fig:cameras}(a) summarize the intrinsic and extrinsic parameters of the stereo event-camera system. The two cameras are hardware-synchronized and arranged with an intersection angle close to 90\degree, providing a well-conditioned geometry for triangulation. Validation with ten scale targets (Fig.~\ref{fig:cameras}(b), Table~\ref{tab:Measurement}) yielded a maximum relative error of 0.56\%, supporting the use of this setup for separation-height measurement.
	
	Fig.~\ref{fig:particle-event} illustrates the stereo event cameras' imaging performance during the particle--surface detachment process. The sequence captures both the instant of lift-off and the subsequent evolution of the detached particle. Using the triangulation procedure in Section~\ref{sec3.4}, the separation height for this example is 15.94~mm, reported with 0.01~mm numeric precision in the current implementation.
	
	Fig.~\ref{fig:data2} further evaluates the temporal evolution of the extracted particle contours. Thanks to the high dynamic range and microsecond temporal precision of the event sensors, the detachment moment is reliably detected even under intense flame luminosity and dense smoke, where frame-based cameras often suffer from saturation or motion blur. The temporal evolution of equivalent particle size from the left and right views shows a synchronized discontinuity at the detected lift-off moment, demonstrating strong internal consistency of the stereo measurements. After separation, the smooth and coherent size trajectories in both views indicate that the combination of SVE-guided event clustering and stereo triangulation (Section~\ref{sec3.4}) enables robust, jump-free particle tracking---crucial for accurately characterizing early post-detachment particle behavior.
	
	\subsection{Cross-Modal Consistency Check of Equivalent Particle Size Measurement}\label{sec4.4}
\begin{figure*}[tp]
		\centering
		\includegraphics[width=0.8\textwidth]{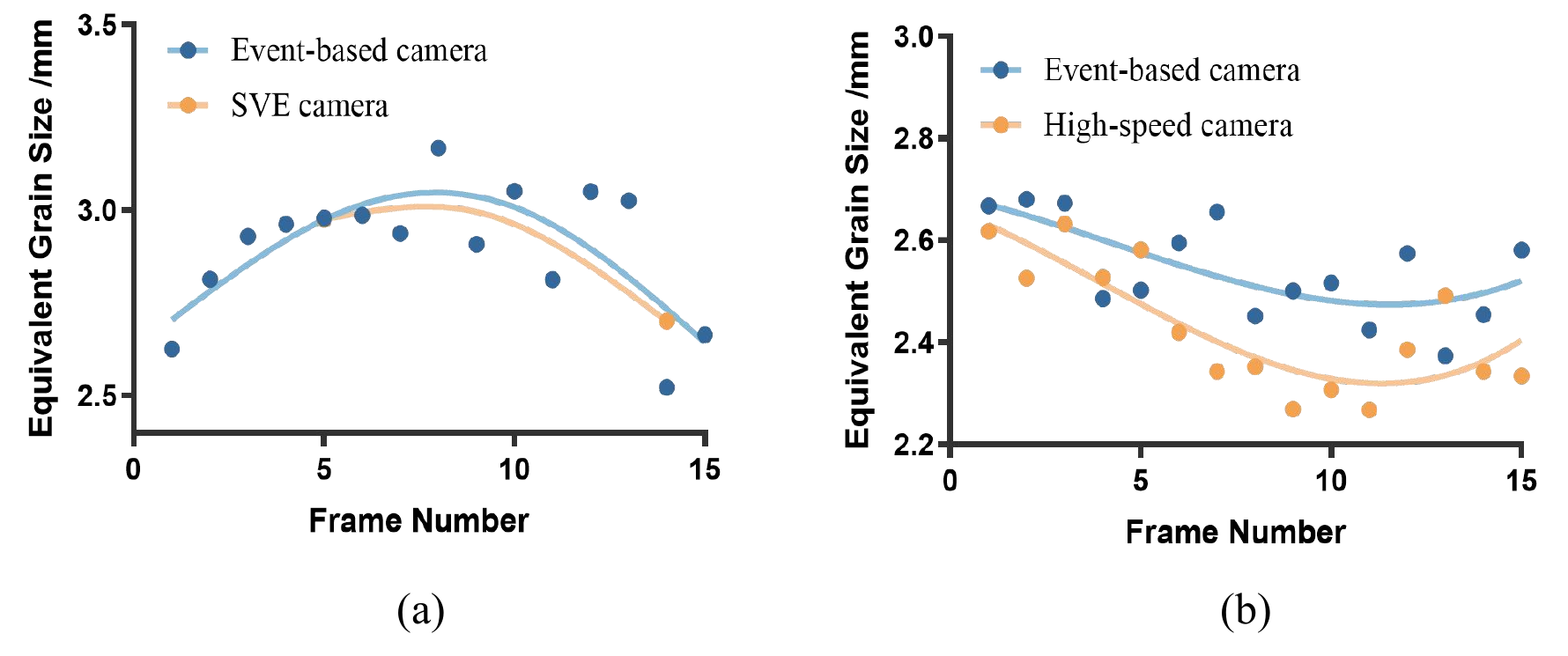}
		\caption{Equivalent particle size measurement results: (a) SVE camera and left event camera; (b) high-speed camera and right event camera.}
		\label{fig:data3}
	\end{figure*}

	Fig.~\ref{fig:data3} presents equivalent particle size measurements obtained from multiple perspectives and modalities. Owing to the inherently irregular and evolving morphology of combustion agglomerates, the apparent particle size depends on both the viewing angle and focal plane position. To evaluate robustness under these practical conditions, the event-based estimates were compared with comparative measurements from the SVE camera and a conventional high-speed camera.
	
	In Fig.~\ref{fig:data3}(a), the size evolution curves obtained from the event camera and SVE images exhibit a closely matched trend over time, despite being generated by independent sensing mechanisms and processing pipelines. This agreement supports the role of the smoke-aware HDR guidance in stabilizing the subsequent event-based contour extraction, and it is presented as cross-modal consistency evidence rather than an absolute ground-truth validation.
	
	Fig.~\ref{fig:data3}(b) compares the event-based estimates with measurements from a high-speed camera. Overall temporal trends are similar; however, the high-speed camera occasionally reports smaller equivalent radii. One plausible factor is its narrower depth of field and greater susceptibility to motion blur: as particles move out of the focal plane or traverse high-luminance regions, their apparent contours become blurred and low-contrast, leading to systematic bias in standard segmentation and fitting. In contrast, the event camera, being less sensitive to motion blur and saturation, consistently provides sharp, high-contrast point clouds for circular fitting, yielding more stable size estimates. The three-way comparison across SVE, event, and high-speed modalities provides converging evidence that the equivalent-size estimates remain consistent under the tested conditions.
	
	\begin{figure*}[tp]   
		\centering
		\includegraphics[width=0.95\textwidth]{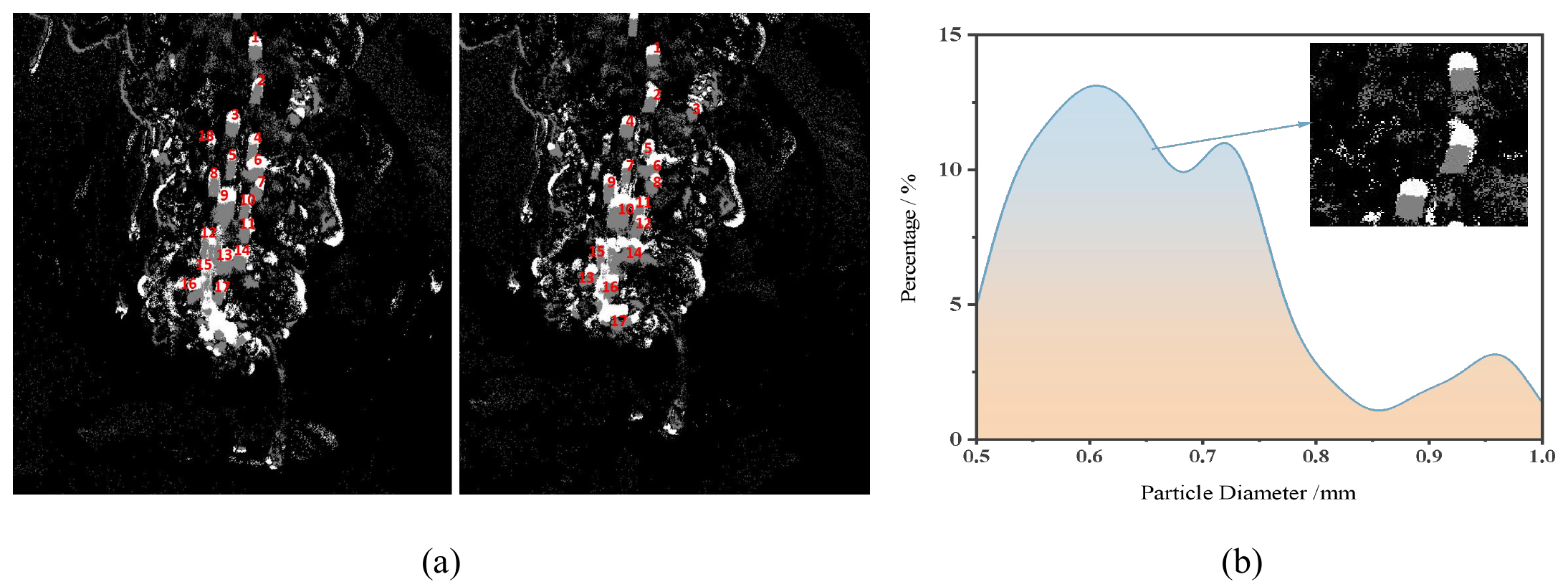}
		\caption{Stereo event-camera particle tracking and size analysis. (a) Stereo event-camera observation of particles and matching effects. (b) Equivalent particle size measurement results.}
		\label{fig:matching}
	\end{figure*} 
	
	\subsection{Statistical Analysis of Size Distributions}\label{sec4.5}
	Beyond single-particle trajectories, the microsecond temporal resolution of event cameras enables frame-level statistics of the equivalent particle size distribution over the combustion process. Fig.~\ref{fig:matching}(a) presents partial particle observation results from the two-view event cameras, where particles are imaged with well-defined contours. By applying time-window constraints, epipolar constraints, and epipolar rectification, corresponding particles are extracted across the two views. The following statistics are computed from particles that can be reliably observed and stereo-confirmed under the given optical-access and visibility conditions.
	
	Fig.~\ref{fig:matching}(b) shows a representative histogram at a specific combustion stage. The distribution exhibits a clear multimodal structure, with several peaks corresponding to dominant size classes. These peaks may reflect different agglomeration and breakup behaviors, although the histogram is limited to stereo-confirmed particles. Due to smoke-induced visibility constraints, detection remains incomplete; the histogram therefore represents the stereo-confirmed population that is reliably observable under the tested conditions.
	
	The inset in Fig.~\ref{fig:matching}(b) shows a typical particle morphology corresponding to the most prominent mode. This morphological evidence supports the interpretation that the most frequently observed equivalent particle size is associated with a characteristic ``coral-like'' aggregate scale, consistent with previous findings on boron-based propellants \cite{Survey,LIU2020105595,Rashkovskiy2017DirectNS}. The ability to resolve such multimodality in real time, thereby eliminating the need for ex-situ sampling or large-scale facilities, highlights a primary strength of neuromorphic sensing guided by SVE priors. This integration of high dynamic range and microsecond resolution allows for the statistically significant observation of both common and outlier events in the evolving particle size spectrum.
	
	\subsection{Discussion, Limitations, and Future Directions}\label{sec4.6}
	The experimental results demonstrate that the proposed SVE--event framework can generate smoke-suppressed HDR images, measure particle separation height, and estimate equivalent particle size with consistent trends. These findings, however, come with several limitations that merit discussion and motivate future work.
	
	First, all experiments were performed using a specific boron-based propellant formulation and a fixed optical setup. Although representative of a practically relevant class of engines, extending the framework to other chemistries or pressure regimes may require adjustments to the SVE transmittance pattern and the smoke-density weighting.
	
	Second, our evaluation relies on verified stereo geometry and cross-modal consistency checks, rather than an independent in-situ reference during combustion. The reported separation height and particle size should therefore be interpreted as internally consistent and physically plausible estimates, not as measurements validated against an external ground truth. Obtaining an independent in-situ reference under smoke-obscured, high-dynamic-range combustion would require a dedicated validation rig and controlled experiments beyond the scope of the present study, which we plan to pursue in future work.
	
	Third, although the present analysis emphasizes separation height and equivalent particle size, the same dataset can also support additional metrological studies, such as 3D trajectory curvature or particle residence time, which offers potential for extending the diagnostic scope.
	
	Fourth, the current implementation operates offline; converting the framework into an online diagnostic tool would require further optimization and potentially hardware acceleration to meet real-time constraints.
	
	Notwithstanding these constraints, the present results demonstrate that coupled SVE--event sensing can recover particle-level diagnostics under conditions that individually challenge each modality. Extending the framework to broader propellant chemistries and operating pressures will require dedicated validation campaigns with independent reference measurements.
	
	\section{Conclusion}\label{sec5}
	This study presents an integrated optical measurement system for high-energy propellant combustion, where extreme dynamic range and ultrafast particle motion occur simultaneously. The system combines an SVE camera with neuromorphic event cameras, achieving high-fidelity imaging and precise quantification of combustion dynamics. The SVE camera provides adaptive HDR guidance to the event cameras, compensating for their lack of absolute intensity and overcoming the limitations of conventional imaging under extreme luminous and transient conditions. 
	
	Experiments on boron-based propellants show that the synergistic framework supports quantitative analysis of key combustion observables under the tested conditions. By combining SVE-guided HDR imaging for feature preservation with event-based stereo reconstruction for microsecond-resolved tracking, we report in-situ measurements of particle separation dynamics and contour-based equivalent-size statistics. The resulting size distributions exhibit multimodal structure in the observed sequences, which is difficult to access with conventional frame-based sensing under similar smoke and saturation conditions. These results illustrate how neuromorphic event sensing, when guided by HDR intensity priors, can advance quantitative propellant diagnostics and inform formulation optimization aimed at suppressing combustion instability.
	
	\section*{Acknowledgements}
	This work was supported by the National Natural Science Foundation of China (Grant No. 12372189).
	
	\section*{Disclosures}
	The authors have no conflicts to disclose.
	
	\section*{Data availability}
	The data that support the findings of this study are available from the corresponding authors upon reasonable request.
	
	\bigskip
	
	\bibliographystyle{unsrtnat}
	\bibliography{sample}

@PREAMBLE{
	"\providecommand{\noopsort}[1]{}" 
	# "\providecommand{\singleletter}[1]{#1}%" 
}

@INPROCEEDINGS{SVE,
	author={Nayar, S.K. and Mitsunaga, T.},
	booktitle={Proceedings IEEE Conference on Computer Vision and Pattern Recognition. CVPR 2000 (Cat. No.PR00662)}, 
	title={High dynamic range imaging: spatially varying pixel exposures}, 
	year={2000},
	volume={1},
	number={},
	pages={472-479 vol.1},
	doi={10.1109/CVPR.2000.855857}}

@article{Bioinspired,
	title={Bioinspired sensors and applications in intelligent robots: a review},
	author={Yanmin Zhou and Zheng Yan and Ye Yang and Zhipeng Wang and Ping Lu and Philip F. Yuan and Bin He},
	journal={Robotic Intelligence and Automation},
	issue={2},
	pages={215-228},
	year={2024},
}

@article{global,
	title={A 240×180 130 dB 3us latency global shutter spatiotemporal vision sensor(Article)},
	author={Brandli, Christian and Berner, Raphael and Yang, Minhao and Liu, Shih-Chii and Delbruck, Tobi},
	journal={IEEE Journal of Solid-State Circuits},
	issue={10},
	volume={49},
	pages={2333-2341},
	year={2014},
}

@article{2009Exposure,
	title={Exposure Fusion: A Simple and Practical Alternative to High Dynamic Range Photography},
	author={ Mertens, Tom  and  Kautz, Jan  and  Reeth, Frank Van },
	journal={Computer Graphics Forum},
	volume={28},
	number={1},
	pages={161-171},
	year={2009},
}

@article{PESPD-MEF,
	title = {Multi-exposure image fusion via perception enhanced structural patch decomposition},
	journal = {Information Fusion},
	volume = {99},
	pages = {101895},
	year = {2023},
	issn = {1566-2535},
	author = {Junchao Zhang and Yidong Luo and Junbin Huang and Ying Liu and Jiayi Ma}
}

@ARTICLE{Simultaneous,
	author={Tao, Jing and Li, You and Guan, Banglei and Shang, Yang and Yu, Qifeng},
	journal={IEEE Transactions on Instrumentation and Measurement}, 
	title={Simultaneous Enhancement and Noise Suppression Under Complex Illumination Conditions}, 
	year={2024},
	volume={73},
	number={},
	pages={1-11}}

@ARTICLE{WGIF,
	author={Kou, Fei and Chen, Weihai and Wen, Changyun and Li, Zhengguo},
	journal={IEEE Transactions on Image Processing}, 
	title={Gradient Domain Guided Image Filtering}, 
	year={2015},
	volume={24},
	number={11},
	pages={4528-4539}}

@article{LIU2020105595,
	title = {Experimental investigation on the condensed combustion products of aluminized GAP-based propellants},
	journal = {Aerospace Science and Technology},
	volume = {97},
	pages = {105595},
	year = {2020},
	issn = {1270-9638},
	doi = {https://doi.org/10.1016/j.ast.2019.105595},
	url = {https://www.sciencedirect.com/science/article/pii/S127096381930313X},
	author = {Huan Liu and Wen Ao and Peijin Liu and Songqi Hu and Xiang Lv and Dongliang Gou and Haiqing Wang}
}

@article{LI2023108126,
	title = {Research on the collision model of high-temperature alumina droplets with cold wall for solid rocket motors},
	journal = {Aerospace Science and Technology},
	volume = {133},
	pages = {108126},
	year = {2023},
	issn = {1270-9638},
	doi = {https://doi.org/10.1016/j.ast.2023.108126},
	url = {https://www.sciencedirect.com/science/article/pii/S1270963823000238},
	author = {Kang Li and Jiang Li and Gen Zhu and Zhipeng He}
}

@article{DEMKO2022112054,
	title = {Observation of aluminum interaction with the binder melt layer using high-speed synchrotron-based phase contrast imaging},
	journal = {Combustion and Flame},
	volume = {241},
	pages = {112054},
	year = {2022},
	issn = {0010-2180},
	doi = {https://doi.org/10.1016/j.combustflame.2022.112054},
	url = {https://www.sciencedirect.com/science/article/pii/S0010218022000736},
	author = {Andrew R. Demko and Kevin J. Hill and Elektra Katz Ismael and Alan Kastengren}
}

@article {PMID:37214717,
	Title = {A Novel Plenoptic Camera-Based Measurement System for the Investigation into Flight and Combustion Behavior of Refuse-Derived Fuel Particles},
	Author = {Zhang, Miao and Vogelbacher, Markus and Aleksandrov, Krasimir and Gehrmann, Hans-Joachim and Stapf, Dieter and Streier, Robin and Wirtz, Siegmar and Scherer, Viktor and Matthes, Jörg},
	DOI = {10.1021/acsomega.2c08004},
	Number = {19},
	Volume = {8},
	Month = {May},
	Year = {2023},
	Journal = {ACS omega},
	ISSN = {2470-1343},
	Pages = {16700—16712}
}

@article{Yang2011ParticleSM,
	title={Particle size measurement of fuel-rich solid propellant using laser attenuation method},
	author={Liu Yang},
	volume = {34},
	pages = {533-536},
	journal={Journal of Solid Rocket Technology},
	year={2011},
	url={https://api.semanticscholar.org/CorpusID:137724537}
}

@article{JIN2020106066,
	title = {Three-dimensional spatial distributions of agglomerated particles on and near the burning surface of aluminized solid propellant using morphological digital in-line holography},
    journal = {Aerospace Science and Technology},
    volume = {106},
    pages = {106066},
    year = {2020},
    issn = {1270-9638},
    doi = {https://doi.org/10.1016/j.ast.2020.106066},
    url = {https://www.sciencedirect.com/science/article/pii/S1270963820307483},
   author = {Bing-ning Jin and Zhi-xin Wang and Geng Xu and Wen Ao and Pei-jin Liu},
}

@ARTICLE{1315998,
	author={Gang Lu and Yong Yan and Colechin, M.},
	journal={IEEE Transactions on Instrumentation and Measurement}, 
	title={A digital imaging based multifunctional flame monitoring system}, 
	year={2004},
	volume={53},
	number={4},
	pages={1152-1158},
	doi={10.1109/TIM.2004.830571}}

@article{LI2024113342,
	title = {New discovery of aluminium agglomeration in composite solid propellants based on microscopic heating system},
	journal = {Combustion and Flame},
	volume = {263},
	pages = {113342},
	year = {2024},
	issn = {0010-2180},
	doi = {https://doi.org/10.1016/j.combustflame.2024.113342},
	url = {https://www.sciencedirect.com/science/article/pii/S001021802400052X},
	author = {Shipo Li and Zhan Wen and Lu Liu and Xiang Lv and Peijin Liu and Bo Yin and Larry K.B. Li and Wen Ao}
}

@ARTICLE{10497132,
	author={Zheng, Zhao and Fan, Jingfan and Shao, Long and Yang, Heqiang and Zhao, Liming and Song, Hong and Ai, Danni and Fu, Tianyu and Xiao, Deqiang and Wang, Yongtian and Yang, Jian},
	journal={IEEE Transactions on Instrumentation and Measurement}, 
	title={Affine Constrained Binocular Camera Calibration With Planar Target Single-Axis Translating}, 
	year={2024},
	volume={73},
	number={},
	pages={1-13},
	doi={10.1109/TIM.2024.3387493}}

@inproceedings{Modeling,
	author = {Verbiest, Frank and Proesmans, Marc and Van Gool, Luc},
	title = {Modeling the Effects of Windshield Refraction for Camera Calibration},
	year = {2020},
	isbn = {978-3-030-58538-9},
	publisher = {Springer-Verlag},
	address = {Berlin, Heidelberg},
	url = {https://doi.org/10.1007/978-3-030-58539-6_24},
	doi = {10.1007/978-3-030-58539-6_24},
	booktitle = {Computer Vision – ECCV 2020: 16th European Conference, Glasgow, UK, August 23–28, 2020, Proceedings, Part VI},
	pages = {397–412},
	numpages = {16},
	location = {Glasgow, United Kingdom}
}

@article{Unsupervised,
	author = {Tonioni, Alessio and Poggi, Matteo and Mattoccia, Stefano and Stefano, Luigi Di},
	title = {Unsupervised Domain Adaptation for Depth Prediction from Images},
	year = {2020},
	issue_date = {Oct. 2020},
	publisher = {IEEE Computer Society},
	address = {USA},
	volume = {42},
	number = {10},
	issn = {0162-8828},
	url = {https://doi.org/10.1109/TPAMI.2019.2940948},
	doi = {10.1109/TPAMI.2019.2940948},
	journal = {IEEE transactions on pattern analysis machine intelligence},
	month = oct,
	pages = {2396–2409},
	numpages = {14}
}

@article{2022Experimental,
	title={Experimental study on the combustion characteristics of aluminized nitrate ester plasticized polyether solid propellant under high pressure},
	author={ Sha, Benshang  and  Na, Xudong  and  Xia, Zhixun  and  Yan, Xiaoting  and  Li, Yang  and  Huang, Liya },
	journal={Acta Astronautica},
	volume={193},
	pages={100-109},
	year={2022},
}

@article{Numerical,
	title = {Numerical simulation of aluminum particle agglomeration near the burning surface of solid propellants},
	journal = {Fuel},
	volume = {342},
	pages = {127767},
	year = {2023},
	issn = {0016-2361},
	doi = {https://doi.org/10.1016/j.fuel.2023.127767},
	url = {https://www.sciencedirect.com/science/article/pii/S0016236123003800},
	author = {Hui Liu and Guangxue Zhang and Jifei Yuan and Zexu Li and Jianzhong Liu},
}

@article{BELAL2018410,
	title = {Ignition and combustion behavior of mechanically activated Al–Mg particles in composite solid propellants},
	journal = {Combustion and Flame},
	volume = {194},
	pages = {410-418},
	year = {2018},
	issn = {0010-2180},
	doi = {https://doi.org/10.1016/j.combustflame.2018.04.010},
	url = {https://www.sciencedirect.com/science/article/pii/S0010218018301627},
	author = {Hatem Belal and Chang W. Han and Ibrahim E. Gunduz and Volkan Ortalan and Steven F. Son},
}

@ARTICLE{niqe,
	author={Mittal, Anish and Soundararajan, Rajiv and Bovik, Alan C.},
	journal={IEEE Signal Processing Letters}, 
	title={Making a “Completely Blind” Image Quality Analyzer}, 
	year={2013},
	volume={20},
	number={3},
	pages={209-212},
	doi={10.1109/LSP.2012.2227726}}

@ARTICLE{niqmc,
	author={Gu, Ke and Lin, Weisi and Zhai, Guangtao and Yang, Xiaokang and Zhang, Wenjun and Chen, Chang Wen},
	journal={IEEE Transactions on Cybernetics}, 
	title={No-Reference Quality Metric of Contrast-Distorted Images Based on Information Maximization}, 
	year={2017},
	volume={47},
	number={12},
	pages={4559-4565},
	doi={10.1109/TCYB.2016.2575544}}

@article{SALGANSKY,
	title = {Effect of solid fuel characteristics on operating conditions of low-temperature gas generator for high-speed flying vehicle},
	journal = {Aerospace Science and Technology},
	volume = {109},
	pages = {106420},
	year = {2021},
	issn = {1270-9638},
	doi = {https://doi.org/10.1016/j.ast.2020.106420},
	url = {https://www.sciencedirect.com/science/article/pii/S1270963820311020},
	author = {E.A. Salgansky and N.A. Lutsenko}
}

@article{PATEL,
	title = {Regression rates and combustion characteristics of dicyclopentadiene based solid fuels with ball milled boron-PTFE additives},
	journal = {Combustion and Flame},
	volume = {275},
	pages = {114035},
	year = {2025},
	issn = {0010-2180},
	doi = {https://doi.org/10.1016/j.combustflame.2025.114035},
	url = {https://www.sciencedirect.com/science/article/pii/S0010218025000732},
	author = {Dhruval N. Patel and Kyle E. Uhlenhake and Justin Kruse and Metin Örnek and Steven F. Son}
}

@article{TAO2025,
	title = {Perceptual region-driven infrared-visible co-fusion for extreme scene enhancement},
	journal = {Optics and Laser Technology},
	volume = {192},
	pages = {113870},
	year = {2025},
	issn = {0030-3992},
	doi = {https://doi.org/10.1016/j.optlastec.2025.113870},
	url = {https://www.sciencedirect.com/science/article/pii/S0030399225014616},
	author = {Jing Tao and Yonghong Zong and Banglei Guan and Pengju Sun and Taihang Lei and Yang Shang and Qifeng Yu}
}

@article{RASHKOVSKIY2019277,
	title = {Formation of solid residues in combustion of boron-containing solid propellants},
	journal = {Acta Astronautica},
	volume = {158},
	pages = {277-285},
	year = {2019},
	issn = {0094-5765},
	doi = {https://doi.org/10.1016/j.actaastro.2019.03.034},
	url = {https://www.sciencedirect.com/science/article/pii/S0094576519304795},
	author = {Sergey A. Rashkovskiy}
}

@article{Rashkovskiy03082017,
	author = {Sergey A. Rashkovskiy},
	title = {Direct Numerical Simulation of Boron Particle Agglomeration in Combustion of Boron-Containing Solid Propellants},
	journal = {Combustion Science and Technology},
	volume = {189},
	number = {8},
	pages = {1277--1293},
	year = {2017},
	publisher = {Taylor and Francis},
	doi = {10.1080/00102202.2017.1294586},
	URL = { https://doi.org/10.1080/00102202.2017.1294586},
	eprint = { https://doi.org/10.1080/00102202.2017.1294586}
	}

@article{Rashkovskiy_2018,
	doi = {10.1088/1742-6596/1009/1/012029},
	url = {https://doi.org/10.1088/1742-6596/1009/1/012029},
	year = {2018},
	month = {apr},
	publisher = {IOP Publishing},
	volume = {1009},
	number = {1},
	pages = {012029},
	author = {Rashkovskiy, S A},
	title = {Boron particle agglomeration and formation of solid residues in combustion of boron-containing solid propellants},
	journal = {Journal of Physics: Conference Series}
}

@INPROCEEDINGS{IROS2022,
		author={Ye, Keyang and Gao, Liuzheng and Guan, Banglei},
		booktitle={IEEE/RSJ International Conference on Intelligent Robots and Systems (IROS)}, 
		title={Visual Odometry in HDR Environments by Using Spatially Varying Exposure Camera}, 
		year={2022},
		volume={},
		number={},
		pages={5995-6000},
		doi={10.1109/IROS47612.2022.9981538}}

@ARTICLE{Solid-State,
			author={Gulve, Rahul and Sarhangnejad, Navid and Dutta, Gairik and Sakr, Motasem and Nguyen, Don and Rangel, Roberto and Chen, Wenzheng and Xia, Zhengfan and Wei, Mian and Gusev, Nikita and Lin, Esther Y. H. and Sun, Xiaonong and Hanxu, Leo and Katic, Nikola and Abdelhadi, Ameer M. S. and Moshovos, Andreas and Kutulakos, Kiriakos N. and Genov, Roman},
			journal={IEEE Journal of Solid-State Circuits}, 
			title={39 000-Subexposures/s Dual-ADC CMOS Image Sensor With Dual-Tap Coded-Exposure Pixels for Single-Shot HDR and 3-D Computational Imaging}, 
			year={2023},
			volume={58},
			number={11},
			pages={3150-3163},
			doi={10.1109/JSSC.2023.3275271}}

@ARTICLE{1683901,
				author={Storm, G. and Henderson, R. and Hurwitz, J.E.D. and Renshaw, D. and Findlater, K. and Purcell, M.},
				journal={IEEE Journal of Solid-State Circuits}, 
				title={Extended Dynamic Range From a Combined Linear-Logarithmic CMOS Image Sensor}, 
				year={2006},
				volume={41},
				number={9},
				pages={2095-2106},
				doi={10.1109/JSSC.2006.880613}}

@article{BOUVIER2014146,
					title = {Logarithmic Image Sensor for Wide Dynamic Range Stereo Vision System},
					journal = {Procedia Computer Science},
					volume = {39},
					pages = {146-149},
					year = {2014},
					issn = {1877-0509},
					doi = {https://doi.org/10.1016/j.procs.2014.11.021},
					url = {https://www.sciencedirect.com/science/article/pii/S1877050914014392},
					author = {Christian Bouvier and Yang Ni}
				}

@INPROCEEDINGS{9746779,
			author={Karakaya, Diclehan and Ulucan, Oguzhan and Turkan, Mehmet},
			booktitle={ICASSP 2022 - 2022 IEEE International Conference on Acoustics, Speech and Signal Processing (ICASSP)}, 
			title={Pas-Mef: Multi-Exposure Image Fusion Based On Principal Component Analysis, Adaptive Well-Exposedness And Saliency Map}, 
			year={2022},
			volume={},
			number={},
			pages={2345-2349},
			doi={10.1109/ICASSP43922.2022.9746779}}

@InProceedings{Jiang_2023_ICCV,
				author    = {Jiang, Ting and Wang, Chuan and Li, Xinpeng and Li, Ru and Fan, Haoqiang and Liu, Shuaicheng},
				title     = {MEFLUT: Unsupervised 1D Lookup Tables for Multi-exposure Image Fusion},
				booktitle = {Proceedings of the IEEE/CVF International Conference on Computer Vision (ICCV)},
				month     = {October},
				year      = {2023},
				pages     = {10542-10551}
			}

@ARTICLE{8839746,
				author={Fang, Yuming and Zhu, Hanwei and Ma, Kede and Wang, Zhou and Li, Shutao},
				journal={IEEE Transactions on Image Processing}, 
				title={Perceptual Evaluation for Multi-Exposure Image Fusion of Dynamic Scenes}, 
				year={2020},
				volume={29},
				number={},
				pages={1127-1138},
				doi={10.1109/TIP.2019.2940678}}

@INPROCEEDINGS{8237767,
					author={Prabhakar, K. Ram and Srikar, V Sai and Babu, R. Venkatesh},
					booktitle={2017 IEEE International Conference on Computer Vision (ICCV)}, 
					title={DeepFuse: A Deep Unsupervised Approach for Exposure Fusion with Extreme Exposure Image Pairs}, 
					year={2017},
					volume={},
					number={},
					pages={4724-4732},
					doi={10.1109/ICCV.2017.505}}

@article{assessment2025,
						title = {An assessment of event-based imaging velocimetry for efficient estimation of low-dimensional coordinates in turbulent flows},
						journal = {Experimental Thermal and Fluid Science},
						volume = {164},
						pages = {111425},
						year = {2025},
						issn = {0894-1777},
						doi = {https://doi.org/10.1016/j.expthermflusci.2025.111425},
						url = {https://www.sciencedirect.com/science/article/pii/S0894177725000196},
						author = {Luca Franceschelli and Christian E. Willert and Marco Raiola and Stefano Discetti}
					}

@article{ams2025,
						author = {Lei Taihang and Guan Banglei and Liang Minzu and Li Xiangyu and Liu Jianbing and Tao Jing and Shang Yang and Yu Qifeng},
						title = {Event-based multi-view photogrammetry for high-dynamic, high-velocity target measurement},
						journal = {Acta Mechanica Sinica},
						year = {2025},
						pages={},
						url = {http://www.sciengine.com/publisher/The Chinese Society of Theoretical and Applied Mechanics/journal/Acta Mechanica Sinica///10.1007/s10409-025-25314-x},
						doi = {https://doi.org/10.1007/s10409-025-25314-x}
					}

@INPROCEEDINGS{8593561,
						author={Liu, Peidong and Geppert, Marcel and Heng, Lionel and Sattler, Torsten and Geiger, Andreas and Pollefeys, Marc},
						booktitle={2018 IEEE/RSJ International Conference on Intelligent Robots and Systems (IROS)}, 
						title={Towards Robust Visual Odometry with a Multi-Camera System}, 
						year={2018},
						volume={},
						number={},
						pages={1154-1161},
						doi={10.1109/IROS.2018.8593561}}

@article{GUO2025111945,
							title = {Structural vibration measurement based on improved phase-based motion magnification and deep learning},
							journal = {Mechanical Systems and Signal Processing},
							volume = {224},
							pages = {111945},
							year = {2025},
							issn = {0888-3270},
							doi = {https://doi.org/10.1016/j.ymssp.2024.111945},
							url = {https://www.sciencedirect.com/science/article/pii/S0888327024008434},
							author = {Liujun Guo and Wenhua Guo and Dingshi Chen and Binxin Duan and Zifan Shi}
						}

@article{D0LC00556H,
						title = {High-speed particle detection and tracking in microfluidic devices using event-based sensing},
						journal = {Lab Chip},
						volume = {20},
						issue = {16},
						pages = {3024-3035},
						year = {2020},
						doi = {10.1039/D0LC00556H},
						url = {https://dx.doi.org/10.1039/D0LC00556H},
						author = {Jessie Howell and Tansy C. Hammarton and Yoann Altmann and Melanie Jimenez},
						publisher = {The Royal Society of Chemistry}
					}

@article{Mueggler2016TheED,
						title={The event-camera dataset and simulator: Event-based data for pose estimation, visual odometry, and SLAM},
						author={Elias Mueggler and Henri Rebecq and Guillermo Gallego and Tobi Delbr{\"u}ck and Davide Scaramuzza},
						journal={The International Journal of Robotics Research},
						year={2016},
						volume={36},
						pages={142 - 149},
						url={https://api.semanticscholar.org/CorpusID:9865213}
					}

@ARTICLE{10301562,
						author={Shiba, Shintaro and Hamann, Friedhelm and Aoki, Yoshimitsu and Gallego, Guillermo},
						journal={IEEE Transactions on Pattern Analysis and Machine Intelligence}, 
						title={Event-Based Background-Oriented Schlieren}, 
						year={2024},
						volume={46},
						number={4},
						pages={2011-2026},
						doi={10.1109/TPAMI.2023.3328188}}

@ARTICLE{Survey,
					author={Gallego, Guillermo and Delbrück, Tobi and Orchard, Garrick and Bartolozzi, Chiara and Taba, Brian and Censi, Andrea and Leutenegger, Stefan and Davison, Andrew J. and Conradt, Jörg and Daniilidis, Kostas and Scaramuzza, Davide},
					journal={IEEE Transactions on Pattern Analysis and Machine Intelligence}, 
					title={Event-Based Vision: A Survey}, 
					year={2022},
					volume={44},
					number={1},
					pages={154-180},
					doi={10.1109/TPAMI.2020.3008413}}

@INPROCEEDINGS{8953722,
						author={Rebecq, Henri and Ranftl, René and Koltun, Vladlen and Scaramuzza, Davide},
						booktitle={IEEE/CVF Conference on Computer Vision and Pattern Recognition (CVPR)}, 
						title={Events-To-Video: Bringing Modern Computer Vision to Event Cameras}, 
						year={2019},
						volume={},
						number={},
						pages={3852-3861},
						doi={10.1109/CVPR.2019.00398}}

@article{ZHUO2024119151,
			title = {3D trajectory, morphology and phase discrimination of burning Al/AP propellant particles with time-resolved astigmatic dual-beam interferometric particle imaging},
			journal = {Powder Technology},
			volume = {433},
			pages = {119151},
			year = {2024},
			issn = {0032-5910},
			doi = {https://doi.org/10.1016/j.powtec.2023.119151},
			url = {https://www.sciencedirect.com/science/article/pii/S0032591023009348},
			author = {Zhu Zhuo and Zhenghui Yang and Yangpeng Liu and Bin Shen and Yang Zhang and Shixi Wu and Dongping Chen and Yingchun Wu and Xuecheng Wu}
		}

@article{LIAO2024113666,
			title = {Combustion and energy performance of multiple aluminum-based alloy particles},
			journal = {Combustion and Flame},
			volume = {269},
			pages = {113666},
			year = {2024},
			issn = {0010-2180},
			doi = {https://doi.org/10.1016/j.combustflame.2024.113666},
			url = {https://www.sciencedirect.com/science/article/pii/S0010218024003754},
			author = {Xueqin Liao and Daolun Liang and Fang Wang and Peini Xie and Yukun Chen and Jianzhong Liu}
		}

@article{WU20214401,
			title = {Particle burning behaviors of Al/AP propellant with high-speed digital off-axis holography},
			journal = {Proceedings of the Combustion Institute},
			volume = {38},
			number = {3},
			pages = {4401-4408},
			year = {2021},
			issn = {1540-7489},
			doi = {https://doi.org/10.1016/j.proci.2020.07.135},
			url = {https://www.sciencedirect.com/science/article/pii/S1540748920305861},
			author = {Yingchun Wu and Zhiming Lin and Zhu Zhuo and Shixi Wu and Chongyang Zhou and Longchao Yao and Wen Ao and Xuecheng Wu and Linghong Chen and Kefa Cen}
		}

@article{CHEN2017225,
			title = {Study of aluminum particle combustion in solid propellant plumes using digital in-line holography and imaging pyrometry},
			journal = {Combustion and Flame},
			volume = {182},
			pages = {225-237},
			year = {2017},
			issn = {0010-2180},
			doi = {https://doi.org/10.1016/j.combustflame.2017.04.016},
			url = {https://www.sciencedirect.com/science/article/pii/S0010218017301505},
			author = {Yi Chen and Daniel R. Guildenbecher and Kathryn N.G. Hoffmeister and Marcia A. Cooper and Howard L. Stauffacher and Michael S. Oliver and Ephraim B. Washburn}
		}

@article{Particle2024,
			title={High Speed Particle Image Velocimetry in a Large Engine Prechamber},
			author={Aravind Ramachandran and Rajat Soni and Markus Roßmann and Marc Klawitter and Clemens Gößnitzer and Jakob Woisetschläger and Anton Tilz and Gerhard Pirker and …Andreas Wimmer},
			journal={Flow, Turbulence and Combustion},
			issue={4},
			pages={1003-1023},
			year={2024},
		}

@article{KALMAN2018144,
			title = {Synchrotron-based measurement of aluminum agglomerates at motor conditions},
			journal = {Combustion and Flame},
			volume = {196},
			pages = {144-146},
			year = {2018},
			issn = {0010-2180},
			doi = {https://doi.org/10.1016/j.combustflame.2018.06.013},
			url = {https://www.sciencedirect.com/science/article/pii/S0010218018302487},
			author = {Joseph Kalman and Andrew R. Demko and Bino Varghese and Katarzyna E. Matusik and Alan L. Kastengren},
		}

@article{ZHANG202477,
			title = {Elaborative collection of condensed combustion products of solid propellants: Towards a real Solid Rocket Motor (SRM) operational environment},
			journal = {Chinese Journal of Aeronautics},
			volume = {37},
			number = {1},
			pages = {77-88},
			year = {2024},
			issn = {1000-9361},
			doi = {https://doi.org/10.1016/j.cja.2023.09.006},
			url = {https://www.sciencedirect.com/science/article/pii/S1000936123003102},
			author = {Wenchao ZHANG and Zhimin FAN and Dongliang GOU and Yao SHU and Peijin LIU and Aimin PANG and Wen AO}
		}

@article{GLOTOV202311,
			title = {Combustion features of boron-based composite solid propellants},
			journal = {Acta Astronautica},
			volume = {204},
			pages = {11-24},
			year = {2023},
			issn = {0094-5765},
			doi = {https://doi.org/10.1016/j.actaastro.2022.12.024},
			url = {https://www.sciencedirect.com/science/article/pii/S0094576522006968},
			author = {O.G. Glotov and V.A. Poryazov and G.S. Surodin and I.V. Sorokin and D.A. Krainov}
		}

@INPROCEEDINGS{1326587,
			author={Malvar, H.S. and Li-wei He and Cutler, R.},
			booktitle={2004 IEEE International Conference on Acoustics, Speech, and Signal Processing}, 
			title={High-quality linear interpolation for demosaicing of Bayer-patterned color images}, 
			year={2004},
			volume={3},
			number={},
			pages={iii-485},
			doi={10.1109/ICASSP.2004.1326587}}

@ARTICLE{DCP,
				author={He Kaiming and Sun Jian and Tang Xiaoou},
				journal={IEEE Transactions on Pattern Analysis and Machine Intelligence}, 
				title={Single Image Haze Removal Using Dark Channel Prior}, 
				year={2011},
				volume={33},
				number={12},
				pages={2341-2353},}

@INPROCEEDINGS{2017ICIVC,
					author={Jackson, Jehoiada and Ariyo, Oluwasanmi and Acheampong, Kingsley and others},
					booktitle={International Conference on Image, Vision and Computing (ICIVC)}, 
					title={Hybrid single image dehazing with bright channel and dark channel priors}, 
					year={2017},
					pages={381-385}}

@article{2019Scene,
	title={Scene Segmentation-Based Luminance Adjustment for Multi-Exposure Image Fusion},
	author={ Kinoshita Yuma  and  Kiya Hitoshi },
	journal={IEEE Transactions on Image Processing}, 
	year={2019},
	volume={28},
	number={8},
	pages={4101-4116},
}

@article{Rashkovskiy2017DirectNS,
	title={Direct Numerical Simulation of Boron Particle Agglomeration in Combustion of Boron-Containing Solid Propellants},
	author={Sergey A. Rashkovskiy},
	journal={Combustion Science and Technology},
	year={2017},
	volume={189},
	pages={1277 - 1293},
	url={https://api.semanticscholar.org/CorpusID:99022629}
}

@INPROCEEDINGS{Zhang,
	author={Zhengyou Zhang},
	booktitle={Proceedings of the Seventh IEEE International Conference on Computer Vision}, 
	title={Flexible camera calibration by viewing a plane from unknown orientations}, 
	year={1999},
	volume={1},
	number={},
	pages={666-673 vol.1},
	doi={10.1109/ICCV.1999.791289}}

@article{Tan2025OptimalPG,
		title={Optimal Pose Guidance for Stereo Calibration in 3D Deformation Measurement},
		author={Dongcai Tan and Shunkun Liang and Bin Li and Banglei Guan and Ang Su and Yuan Lin and Dapeng Zhang and Minggang Wan and Zibin Liu and Chenglong Wang and Jiajian Zhu and Zhang Li and Yang Shang and Qifeng Yu},
		journal={Experimental Mechanics},
		year={2025},
		volume={abs/2511.18317},
		doi={https://doi.org/10.1007/s11340-025-01255-1}
	}

@article{guan2026fusion,
		title={Fusion-Restoration Image Processing Algorithm to Improve the High-Temperature Deformation Measurement},
		author={Guan, Banglei and Tan, Dongcai and Tao, Jing and Su, Ang and Shang, Yang and Yu, Qifeng},
		journal={Experimental Mechanics},
		pages={1--12},
		year={2026},
		publisher={Springer}
	}

@article{guan2026,
		title={A DMD-Based Adaptive Modulation Method for High Dynamic Range Imaging in High-Glare Environments}, 
		author={Banglei Guan and Jing Tao and Liang Xu and Dongcai Tan and Pengju Sun and Jianbing Liu and Yang Shang and Qifeng Yu},
		year={2026},
		eprint={2602.12044},
		journal={arXiv},
		primaryClass={cs.CV},
		url={https://arxiv.org/abs/2602.12044}, 
	}
	
\end{document}